\documentclass[conference]{IEEEtran}
\IEEEoverridecommandlockouts
\usepackage{cite}
\usepackage{amsmath,amssymb,amsfonts}
\usepackage{algorithmic}
\usepackage{hyperref}
\usepackage[T1]{fontenc}
\usepackage{times} 
\usepackage{subcaption}

\usepackage{graphicx}
\usepackage{textcomp}
\usepackage{multicol}
\usepackage{xcolor}
\usepackage{marvosym}
\usepackage{float} % For [H] positioning
\usepackage{listings}
\usepackage{amsmath}
\usepackage{booktabs}
\usepackage{multirow}
\usepackage[x11names]{xcolor} % 扩展颜色支持
\usepackage{tcolorbox}
\usepackage{tabularx}  % 自适应宽度表格
\usepackage{cuted}
\usepackage{adjustbox}
\usepackage{array}
\tcbuselibrary{breakable, listings, skins}
\def\BibTeX{{\rm B\kern-.05em{\sc i\kern-.025em b}\kern-.08em
    T\kern-.1667em\lower.7ex\hbox{E}\kern-.125emX}}

\newtcolorbox{mutationbox}{
  title=\textbf{Mutation prompt for TSP},
  colback=white,
  colframe=black,
  colbacktitle=white,
  coltitle=black,
  fonttitle=\bfseries,
  boxrule=0.8pt,
  breakable,                    % 允许跨栏分页
  enhanced,                     % 启用高级功能
  before upper={\parindent0pt}, % 取消首行缩进
  top=8pt,                      % 上边距
  left=4pt,                     % 左边距
  right=4pt,                    % 右边距
  attach boxed title to top left={xshift=5mm,yshift=-2mm}
}
\newtcolorbox{mutationbox_eda}{
  title=\textbf{Mutation prompt for EDA},
  colback=white,
  colframe=black,
  colbacktitle=white,
  coltitle=black,
  fonttitle=\bfseries,
  boxrule=0.8pt,
  breakable,                    % 允许跨栏分页
  enhanced,                     % 启用高级功能
  before upper={\parindent0pt}, % 取消首行缩进
  top=8pt,                      % 上边距
  left=4pt,                     % 左边距
  right=4pt,                    % 右边距
  attach boxed title to top left={xshift=5mm,yshift=-2mm}
}
\newtcolorbox{crossoverbox}{
  title=\textbf{Crossover prompt for TSP},
  colback=white,
  colframe=black,
  colbacktitle=white,
  coltitle=black,
  fonttitle=\bfseries,
  boxrule=0.8pt,
  breakable,                    % 允许跨栏分页
  enhanced,                     % 启用高级功能
  before upper={\parindent0pt}, % 取消首行缩进
  top=8pt,                      % 上边距
  left=4pt,                     % 左边距
  right=4pt,                    % 右边距
  attach boxed title to top left={xshift=5mm,yshift=-2mm}
}
\newtcolorbox{crossoverbox_eda}{
  title=\textbf{Crossover prompt for EDA},
  colback=white,
  colframe=black,
  colbacktitle=white,
  coltitle=black,
  fonttitle=\bfseries,
  boxrule=0.8pt,
  breakable,                    % 允许跨栏分页
  enhanced,                     % 启用高级功能
  before upper={\parindent0pt}, % 取消首行缩进
  top=8pt,                      % 上边距
  left=4pt,                     % 左边距
  right=4pt,                    % 右边距
  attach boxed title to top left={xshift=5mm,yshift=-2mm}
}
\begin{document}

\title{Using Multi-modal Large Language Model to Boost Fireworks Algorithm's Ability in Settling Challenging Optimization Tasks}
% \author{\IEEEauthorblockN{Anonymous Authors}}
\author{\IEEEauthorblockN{Shipeng Cen}
\IEEEauthorblockA{\textit{School of Intelligence Science and Technology, }\\
\textit{Institute for Artificial Intellignce,} \\
 Peking University, Beijing, China\\
censhipeng@pku.edu.cn}

\and
\IEEEauthorblockN{Ying Tan \textsuperscript{\Letter}}
\IEEEauthorblockA{\textit{School of Intelligence Science and Technology,}\\
\textit{Institute for Artificial Intellignce,} \\
 \textit{National Key Laboratory of General Artificial Intelligence,}\\
 Peking University, Beijing, China \\
ytan@pku.edu.cn}}

\maketitle

\begin{abstract}
As optimization problems grow increasingly complex and diverse, advancements in optimization techniques and paradigm innovations hold significant importance. The challenges posed by optimization problems are primarily manifested in their non-convexity, high-dimensionality, black-box nature, and other unfavorable characteristics. Traditional zero-order or first-order methods, which are often characterized by low efficiency, inaccurate gradient information, and insufficient utilization of optimization information, are ill-equipped to address these challenges effectively. In recent years, the rapid development of large language models (LLM) has led to substantial improvements in their language understanding and code generation capabilities. Consequently, the design of optimization algorithms leveraging large language models has garnered increasing attention from researchers. In this study, we choose the fireworks algorithm(FWA) as the basic optimizer and propose a novel approach to assist the design of the FWA by incorporating multi-modal large language model(MLLM). To put it simply, we propose the concept of Critical Part(CP), which extends FWA to complex high-dimensional tasks, and further utilizes the information in the optimization process with the help of the multi-modal characteristics of large language models. We focus on two specific tasks: the \textit{traveling salesman problem }(TSP) and  \textit{electronic design automation problem} (EDA). The experimental results show that FWAs generated under our new framework have achieved or surpassed SOTA results on many problem instances. So our framework shows great potential in addressing NP-hard problems. Meanwhile, we compare and analyze the algorithmic design behavior of multi-modal large language models in different scenarios, and study the impact of visual modal optimization information on the proposed framework, which provides a reference for the research in this field.
\end{abstract}


\section{Introduction}
In the contemporary era of rapid technological advancement, optimization problems have become ubiquitous across various fields, including engineering, finance, logistics, and computer science. These problems are increasingly complex and diverse, often characterized by non-convexity, high-dimensionality, multiple saddle points, and black-box nature. Such characteristics make it extremely challenging to find effective solutions using traditional optimization methods. The "No Free Lunch"\cite{1} theorem further complicates the situation, implying that no single optimization algorithm can universally outperform all others across all possible problems. This means that selecting the most suitable optimization algorithm from the vast array of available options is highly problem-dependent and often requires extensive empirical experience. However, even with experience, it is difficult to ensure that the chosen algorithm is truly optimal for a given problem, leading to potential inefficiencies and suboptimal solutions.

Given these challenges, there is an urgent need for new approaches that can effectively and adaptively address the complexities of modern optimization problems. And the approaches can make full use of optimization information to improve algorithm performance.

 FWA\cite{2} is chosen as the fundamental optimizer. As a swarm intelligence optimization algorithm, the FWA is characterized by its simplicity and comprehensibility in design principles. The performance of the FWA has been extensively validated across various optimization scenarios. It is even capable of facilitating zero-order LoRA fine-tuning of large language models themselves, without the need for gradient information\cite{3}, which also makes the FWA particularly well-suited for optimization tasks that are non-convex and black-box in nature. Moreover, FWA possesses a broad design space at the operator level, including explosion operator, mutation operator and selection operator, which provides a fertile ground for us if we want to customize the optimizer for optimization tasks even without gradient. 

A substantial body of existing research has already demonstrated the advantages of LLM-assisted optimizers design, achieving promising results. Moreover, some scholars have explored the integration of LLMs and FWA\cite{26,28}. However, the current mainstream paradigm remains confined to text-based interactions with large language models, which is insufficient for fully utilizing historical optimization information and optimization results. Also, Previous paradigms have not expanded the scope of FWA applications, such as high-dimensional complex problems.

To address these limitations, we propose a novel paradigm in which various visualizations of the optimization process can be fed to the MLLM to give it a multi-perspective view of the task. And we propose a new concept called critical part(CP), the object that MLLM needs to design. The whole pipline in shown in Fig.\ref{method}). 

We validate the effectiveness and great potential of this framework through experiments on highly challenging NP-hard problems, TSP and EDA tasks. The dimensionality of problems has expanded dramatically, ranging from hundreds to millions of dimensions. The results demonstrate the significant advantages of our proposed approach.

This paper's contributions are as follows,
\begin{itemize}
    \item \textbf{At the framework level.} We improve the previous LLM-driven FWA evolutionary framework by introducing a multimodal large language model in which visual information can be exploited during the optimization process, and by proposing the concept of critical part(CP), which allows the design object to be freely switched between global FWA and critical local FWA, extending the range of problems that can be solved by FWA to complex and high-dimensional tasks such as EDA..
    \item \textbf{At the experiment results level.}  On the TSP task, the algorithm generated based on the new framework achieves the best results compared to other algorithms on numerous TSP instances with known optimal paths in TSPLIB, and it is worth mentioning that some of the paths obtained with visual information are better than those given by TSPLIB in the floating-point sense although they are not optimal solutions in the rounding sense; On the EDA task, relying on the open-source Dreamplace framework, the FWA generated based on the new framework is able to better adjust the built-in optimizer step size, and with a very extremely low computational resource requirement (a 12GB GPU is sufficient), six out of the eight instances have generated better solutions than the SOTA from EOAGP\cite{11}.
    \item \textbf{At the MLLM level.} Starting from the logical similarity of the optimal fireworks algorithm of different examples, we deeply analyze the behavior characteristics of the MLLM in the new framework in the design process of the fireworks algorithm. We draw two key conclusions: (1) The introduction of visual modalities in the optimization process does not guarantee to improve the performance of the overall framework, which contradicts the intuition that "the more information, the stronger the optimization ability". However, for some difficult optimization problems, it is very worth trying to introduce visual modal information (refer to Table II for examples b3 and b4). (2) The introduction of visual modal optimization information affects the algorithm design behavior of MLLM, and if the optimization information of visual modalities can introduce additional "operational" optimization cues, MLLM tends to generate FWA codes with high heterogeneity and customization, specifically refer to the experiment of TSP problem; However, if there is little difference in the optimization information of the visual modal corresponding to different problem instances, the MLLM will generate FWA codes with high similarity, which will refer to the experiment of EDA problem.
\end{itemize}

The structure of this paper is outlined as follows: Section II introduces related work, including the fireworks algorithm and large language model for algorithm design. In Section III, we introduce our method, as depicted in Fig.\ref{method}. In Section IV, we conduct extensive experimental designs, comparisons, and analysis. Finally, Section V concludes the paper.

\section{Related works}
\subsection{Fireworks Algorithm}
Fireworks algorithm is inspired by the behavior of fireworks explosion in the night sky was proposed. As a swarm intelligence optimization algorithm, it has a very clear logical structure, and the overall process includes classical operators such as \textbf{explosion}, \textbf{mutation}, and \textbf{selection}. The traditional paradigm of fireworks algorithm design requires researchers to put great efforts to improve all kinds of operators according to the nature of the problem to improve the task adaptability. This traditional paradigm has achieved excellent results and has been applied in various fields, such as optimization of networks of reinforcement-learning intelligences, efficient parameter fine-tuning of large language models\cite{3}, large-scale knapsack problems\cite{27}, and so on.
%这里加一些引用

In recent years, algorithmic researchers have noticed the great potential of LLM-assisted optimization algorithm design due to the rapid development of large language models. For swarm intelligence optimization algorithms, which benefit from clear optimization logic at the operator level, they are naturally suited to LLM for design. For fireworks algorithms, some work has been done to assist in the design of fireworks algorithms, Cen et al.\cite{26} design the operators of fireworks algorithms and propose a new operator called "novel operator". With the help of LLM, the proposed "novel operator" can better integrate the whole situation, and achieve SOTA results in complex engineering problems with strong constraints. Liu et al.\cite{28} focus on the design of mutation operator for FWA, and compare the results of different large language models, and analyze the process of generating operators in depth, and study the relationship between the length of the code and the performance.

The design paradigm for fireworks algorithms is quietly and rapidly shifting, but we believe that the current paradigm is still lacking in areas, especially in (1) \textbf{insufficient use} of information in the optimization process (2) \textbf{insufficient exploitation} of large language models (3) \textbf{insufficient exploration }of large-scale optimization problems. We will discuss this further in later sections.
\subsection{Large Language Model for Algorithm Design}

In addition to optimization areas such as fireworks algorithms, large language model-assisted heuristic rule design has yielded very impressive results. For example, the early \textbf{Funsearch}\cite{7} from deepmind, as a pioneering work, proved the potential of LLM in code design, improving the solution of a mathematical problem at that time; later, \textbf{EOH}\cite{4}, as a complete original work, not only greatly reduced the consumption of Funsearch by refining the cue words and introducing design ideas, but also greatly improved the task performance, providing a very good reference for subsequent work; the recent presentation of \textbf{Alphaevolve}\cite{6} from deepmind has opened up researchers' horizons even more, with more open mathematical problems systematically investigated and improved in this whitepaper; it also optimizes the underlying design of the large language model behind it, and has the potential to realize the mutual cyclic facilitation of LLMs and algorithms. As the first work to introduce model fine-tuning within the field of algorithmic automatic design, \textbf{CALM}\cite{5} further fine-tunes the LLM based on the results of generated code as a reward signal during the process of algorithmic automatic design, and experimentally demonstrates that RL enhances the heuristic rule-designing capability of the LLM. Meanwhile, CALM makes some improvements to the previous algorithm design framework, such as cue words for algorithm iteration. Another work, \textbf{EOAGP}\cite{11}, uses large language models to help improve DreamPlace\cite{12,13}, an open-source framework in the field of EDA. The whole pipeline includes multi-stage code design (including initialization, preconditioner, and optimizer), costs large computing resources and achieved good results.

As seen above, in the field of automatic algorithm design, LLM is playing a more and more important role. However, as we mentioned before, there are some shortcomings in the research on fireworks algorithms incorporating LLMs, again common to current research. First, previous work has pointed out that using the multi-modal capabilities of large language models can enhance the ability of the large language models themselves to solve mathematical problems, which provides confidence in using multi-modal large language models(MLLM)\cite{14} to aid in algorithm design. From the latest literature now, using  multi-modal large language models to aid algorithm design are still in a rather preliminary stage, and we need to fill the gap as soon as possible. MLLM should better understand task information with visual information, which is very important for our framework.
\section{Methods}

\begin{figure*}[t] % [t]顶部/[b]底部/[htbp]浮动位置
  \centering
  \includegraphics[width=1.0\textwidth]{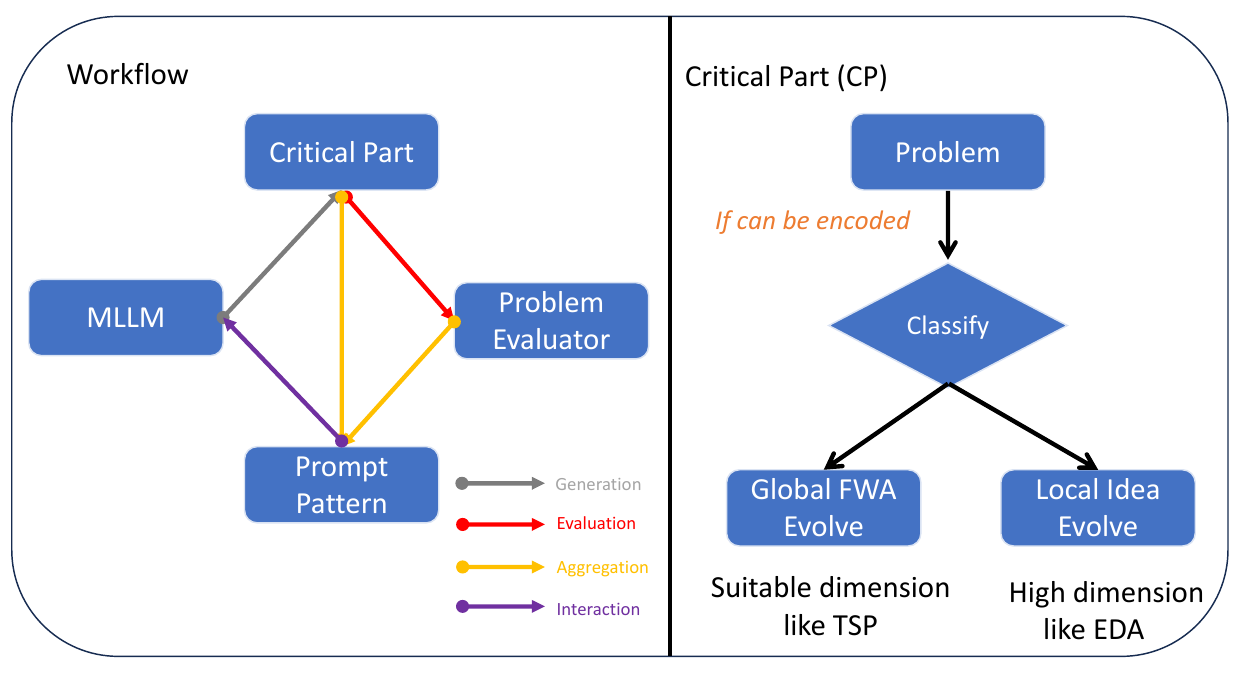} % 调整宽度
  \caption{Left: the overflow of the evolution of Critical Part under the help of Multi-modal large language model;Right: the introduction of CP by classifying the problems}
  \label{method}
\end{figure*}

%先介绍当前的两种范式
In LLM-based algorithm design area, given an optimization task, prior work generally follows two approaches. (1) Introducing dedicated optimizers like FWA by encoding the objectives and constraints into mathematical formulations, then processing them by improving optimizers through traditional workflow (left panel of Fig.\ref{method}), just the paradigm successfully adopted by researchers to solve complex engineering problems\cite{26,28}. (2) Discovering heuristic rules applicable to the optimization task, such as the design principles for underlying large language model architectures referenced in AlphaEvolve or optimizer's step-size design rules in EDA to be discussed later. Typically, these two approaches can operate in parallel. Take the TSP, it can be reformulated as a heuristic rule design challenge focused on iteratively selecting the next city to form an optimal cycle, a strategy proven effective in EOH work. Alternatively, for tasks without encoding difficulties, traditional approaches like discrete FWA can be employed as global optimizers.

The comparative effectiveness of these two paradigms remains an open question. Broadly speaking, heuristic-based algorithm design emphasizes textual interaction with Large Language Models (LLMs), requiring natural language descriptions to explicitly convey constraints, requirements, and objectives to the model. In contrast, optimizer-based algorithm design still necessitates manual upfront problem encoding (including variables, objectives, and constraints). Subsequent iterations, however, can proceed with minimal task descriptions by leveraging a code + performance feedback loop, enabling the LLM to iteratively refine the algorithm.

\begin{figure*}[t] % [t]顶部/[b]底部/[htbp]浮动位置
  \centering
  \includegraphics[width=1.0\textwidth]{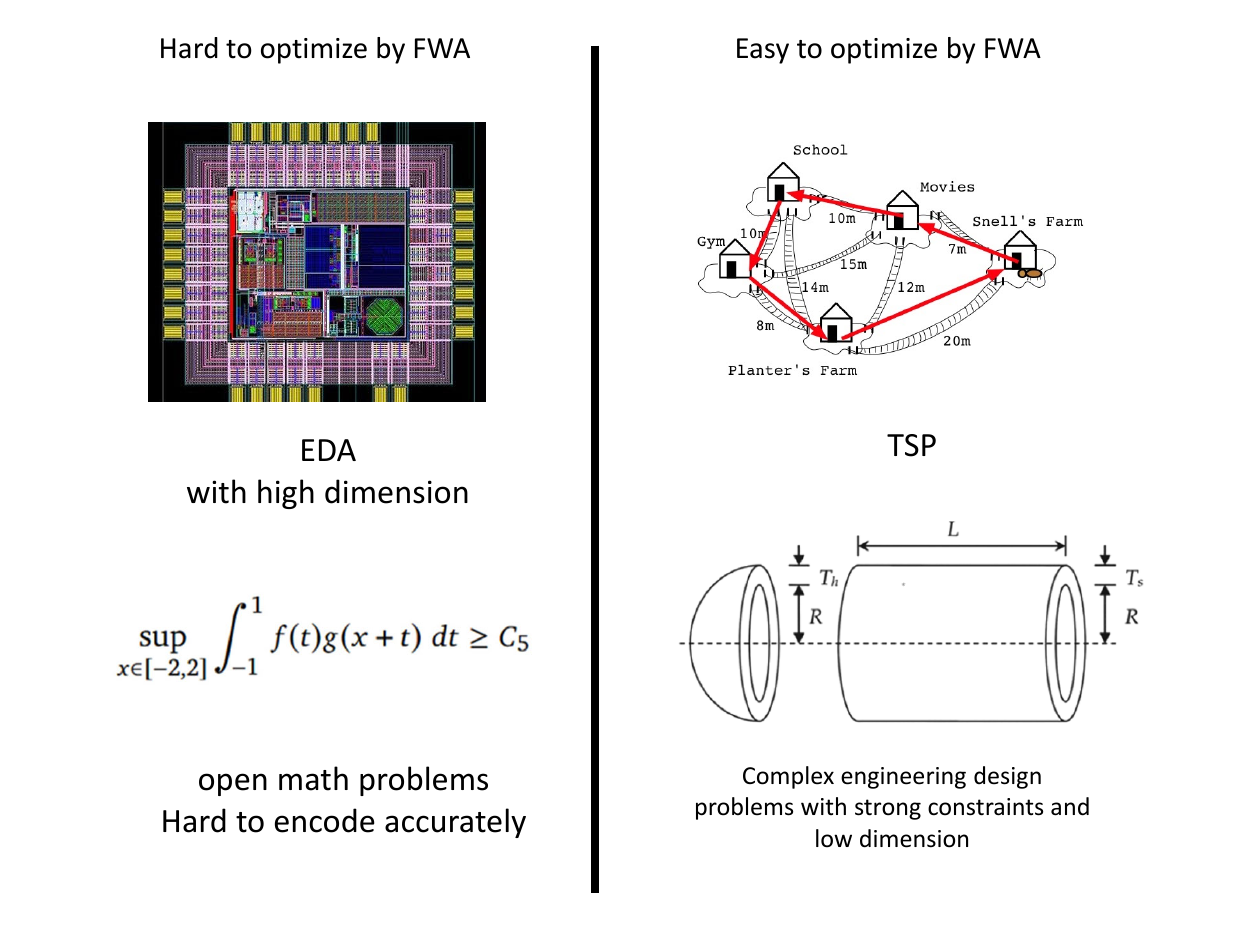} % 调整宽度
  \caption{Classification of problems according to the difficulty of optimization using FWA. \textbf{Left}: EDA tasks are large-scale and not suitable for swarm intelligence algorithms; Open math problem is mostly difficult to encode. \textbf{Right}: Typical problems like TSP and engineer design problems and be solved by FWA}
  \label{task4FWA}
\end{figure*}

%对任务进行分类
To further enhance the LLM-based optimizer design framework capabilities, we first classify FWA applicability as shown in Fig.\ref{task4FWA}. Two representative cases demonstrating FWA limitations include: (1) Problems with very high dimension like EDA, where although solutions permit straightforward optimizer encoding, complex constraints combined with high-dimensional landscapes (frequently millions of variables riddled with saddle points) render purely FWA inadequate; and (2) Problems hard to encode accurately, especially open math problems like Erdős' minimum overlap problem, which can not be perfectly encoded becasue real-valued functions with continuous domains and ranges can not be directly encoded. While approximation techniques exist (e.g., step-function discretization), increasing precision demands cause rapidly dimensionality growth. Though the AlphaEvolve framework achieved record improvements on this problem, progress remains modest and fundamentally constrained by dependency on existing mathematicians' heuristic constructions.

To extend the task-solving capabilities of LLM-FWA, we introduce the Critical Part (CP) illustrated in Fig.\ref{task4FWA} (right), which expands LLM driven FWA's optimization target space by classifying the problem. This implementation operates through two distinct pathways: (1) For optimization problems amenable to global techniques, the CP module directly deploys FWA as the global optimizer; (2) For cases exhibiting high dimensionality and challenging optimization landscapes exemplified by EDA, we forgo global FWA optimization. Instead, we evolve local FWA to refine step-size rule within critical optimizers of the open-source placement framework Dreamplace, whose performance shows extreme sensitivity to optimizer selection and parameterization, with conventional methods like Adam consistently demonstrating suboptimal outcomes.

Now, we will introduce the complete LLM-FWA framework based on the Multimodal Large Language Model. The overall method is shown in Fig.\ref{method}. 

On the left side is the overall workflow including four objects, which are the MLLM, the Prompt Pattern, the Problem Evaluator, and the Critical Part. The process on the left is similar to the previous automated algorithm design process. Every time, we will use different prompt patterns to \textbf{interact} with MLLM. In this work, we mainly use "mutation" and "crossover" prompts to design and improve the set of optimizers. The prompts will also contain the code of the current individual (write the design idea in the comment) and the resulting representation (both in text and visual form).  Then the MLLM  \textbf{generates} and returns the new code of the critical part, which will be sent to the problem evaluator for \textbf{evaluation}. Next, the code of the critical part and the result of the evaluation (both in text and visual form) will be \textbf{aggregated} and recorded for the next round.

On the right is a explanation of the Critical Part. We have already explained its motivation and purpose. More specifically, when we are faced with an optimization problem, we need to judge whether the problem can be encoded into a form suitable for the optimizer, such as the general infinite-dimensional function space, etc., which is not discussed for the time being. If a problem can be coded, researchers need to take the initiative to classify the problem, which mainly relies on personal experience (1) some classic scheduling and planning problems, such as TSP, backpack problems, etc., can directly take the form of optimizers (2) large-scale complex professional problems such as EDA, which need to rely on the existing open source framework, use FWA to improve the underlying optimizer, and it is recommended to take the form of local idea evolution.

\section{Experiments and Discuss}
In this section, our experiments are divided into two parts. (1) we conduct tests on the TSP instances. (2) we conduct tests on the EDA instances. 

Doubao-1.5-pro-vision is used as our multi-modal large language model and we will control whether or not visual optimization information is provided in the experiment for further analysis. Inside these two tasks, we have a pool with population  capacity of 5 managed via greedy selection and a maximum number of iteration of 200, with one mutation and one crossover operation per iteration. Mutation and crossover are attempted up to five times, with no further attempts if they are successful.

\subsection{Traveling Salesman Problem}
TSP is a classic combinatorial optimization problem in computer science and operations research. It belongs to the class of NP-hard problems, making it computationally challenging to solve for large instances. The TSP seeks to determine the shortest possible route that allows a salesman to visit a set of cities exactly once and return to the starting city. It has widespread applications in logistics, route planning, circuit design, and DNA sequencing.

Formally, the problem is defined over a set of cities $V=\{v_1, v_2,...,v_n\}$, with pairwise distance $d_{ij}$ between cities $v_i$ and $v_j$. The goal is to find a \textbf{permutation} $\pi=(\pi_1,\pi_2,...,\pi_n)$ of the cities that minimize the following travel distance:
\[\text{Travel distance} = \sum_{k=1}^{n}d_{\pi_k\pi_{k+1}}, \text{where}\ \pi_{n+1}=\pi_{1} \] 

TSPLIB\cite{15} is the \textbf{authoritative benchmark library} for the Traveling Salesman Problem (TSP), widely adopted for rigorous algorithm evaluation. Maintained by combinatorial optimization experts, it provides standardized test instances with mathematically validated optimal solutions. It contains 50+ instances and the instance sizes range from 17 to 85,900 cities.

The TSP inherently aligns with heuristic optimization algorithms due to its NP-hard combinatorial nature. We adopt the FWA as the global optimizer, where each individual in the population encodes a unique permutation of city visitation sequences. During each iteration, two core evolutionary operators are executed:

\begin{itemize}
    \item \textbf{Mutation} : A candidate individual is randomly selected from the current population. The LLM is tasked with redesigning one of three critical operators – explosion operator, mutation operator, or selection operator – based on contextual analysis of optimization dynamics.
    \item \textbf{Crossover} : Two parent individuals are randomly paired, prompting the LLM to strategically select one operator from each parent (e.g., explosion operator from Parent A and mutation operator from Parent B) and synthesize them through logical composition rules.
\end{itemize}

To enable comprehensive state awareness, each operation feeds diverse inputs (with visual information in multi-modal environment) to the LLM, comprising:

\begin{itemize}
    \item \textbf{Textual information} FWA codes and tour lengths.
    \item \textbf{Visual information} route with its convex hull, path crossing heatmap, and route density.
\end{itemize}

\begin{figure*} % 星号用于创建跨双栏图像
  \centering
  \begin{subfigure}{0.31\textwidth}
    \includegraphics[width=\textwidth]{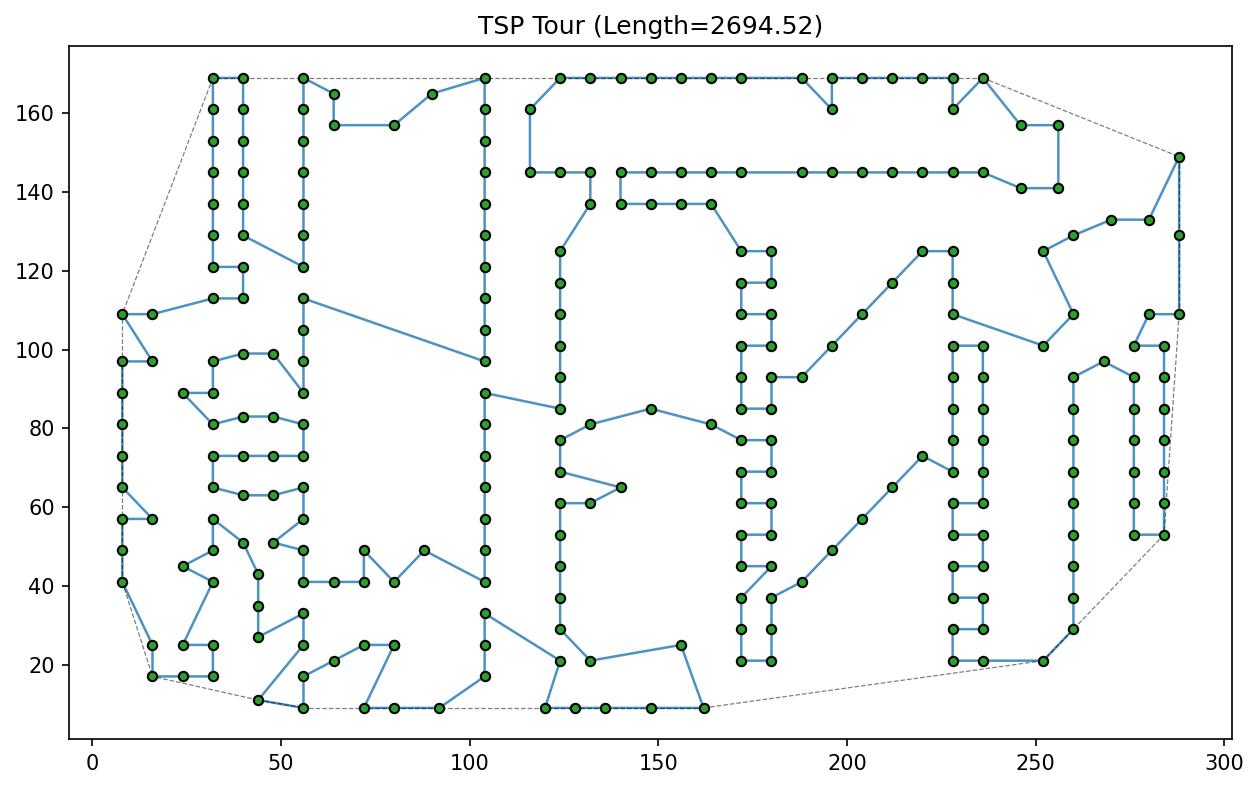}
    \caption{}
    \label{fig:a}
  \end{subfigure}
  \hfill
  \begin{subfigure}{0.31\textwidth}
    \includegraphics[width=\textwidth]{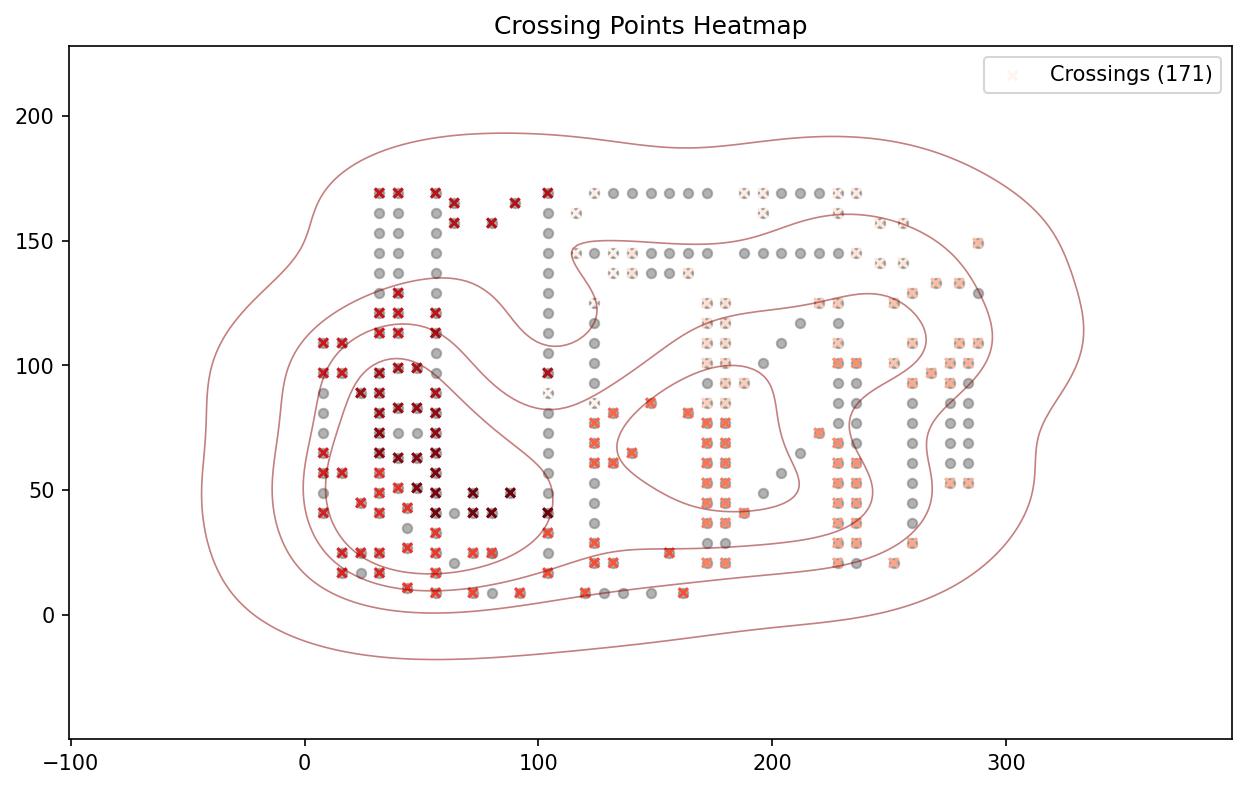}
    \caption{}
    \label{fig:b}
  \end{subfigure}
  \hfill
  \begin{subfigure}{0.31\textwidth}
    \includegraphics[width=\textwidth]{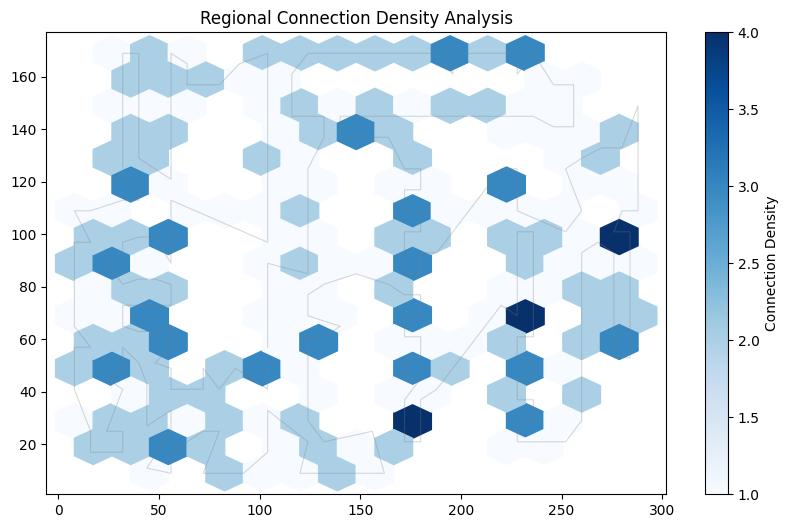}
    \caption{}
    \label{fig:c}
  \end{subfigure}
  
  \caption{Three types of visual information: (a) Path visualization, 
           (b) Path crossing heatmap, (c) Route density analysis}
  \label{fig:triple}
\end{figure*}

The construction of the three visualization information is done as follows:
(1) The visualization of routes and their convex hulls offers straightforward implementation, enabling MLLM to intuitively and qualitatively understand path configurations.
(2) The path crossing heatmap identifies solution deficiencies by computationally determining intersections between paths and visually mapping these crossing points through a heat gradient representation.
(3) The route density analysis calculates midpoints along each path segment, then partitions the plane into regular hexagons to count midpoint densities within each cell. This method effectively visualizes spatial connection distribution patterns, proving particularly valuable for identifying critical regions within the path network. 
You can refer to Fig.\ref{fig:triple} for details.

The overall prompt is as follows:
\begin{mutationbox}
\begin{lstlisting}
You're a specialist in evolutionary algorithms and combinatorial optimization. Please perform mutation on the following high-performing TSP algorithm:
Current Code: {code}
Current Average Distance: {distance}
Current TSP visual information: Please refer to the 3 images I send you, corresponding to basic paths, path cross-analysis and density analysis respectively (Only provided for when visual information is given)

Modification Requirements:
1. Modify ONLY ONE class function in explosion_operator/ mutation_operator/selection_operator, and the code shoud be longer and more powerful. Other class functions' modifications are prohibited.
2. Maintain all functions' original input/output formats
3. Return ONLY the complete correct Python code with necessary imports, no explanations
\end{lstlisting}
\end{mutationbox}

\begin{crossoverbox}
\begin{lstlisting}
You're a specialist in evolutionary algorithms and combinatorial optimization. Please perform crossover on the following two high-performing TSP algorithms:
Parent 1 Code: {{code}}
Parent 1 Distance: {{distance}}
Parent 1 TSP visual information: Please refer to the first 3 images I send you, corresponding to basic paths, path cross-analysis and density analysis respectively(Only provided for when visual information is given)
Parent 2 Code: {{code}}
Parent 2 Distance: {{distance}}
Parent 2 TSP visual information: Please refer to the last 3 images I send you, corresponding to basic paths, path cross-analysis and density analysis respectively(Only provided for when visual information is given)

Modification Requirements:
1. Select one operator from each parent algorithm (explosion_operator/mutation_operator/selection_operator) and combine them logically, and the code shoud be longer and more powerful. Other class functions' modifications are prohibited.
2. Ensure compatibility between the integrated components
3. Maintain all functions' original input/output formats
4. Return ONLY the complete correct Python code with necessary imports, no explanations

Provide the synthesized algorithm that inherits the best characteristics from both parents
\end{lstlisting}
\end{crossoverbox}

\begin{table*}[htbp]
\centering
\caption{Traveling salesman problem results. Comparison of the relative distance (\%) to the best-known solutions (lower is better) for various routing heuristics on a subset of TSPLib instances.}
\label{table:tsp_results}
\resizebox{0,8\textwidth}{!}{
\begin{tabular}{lcccccccc}
\toprule
Algorithm & \multicolumn{8}{c}{TSPLib Instances} \\
\cmidrule(lr){2-9}
 & \textbf{pr76} & \textbf{kroC100} & \textbf{kroD100} & \textbf{kroA100} & \textbf{rd100} & \textbf{eil101} & \textbf{eil51} & \textbf{st70} \\
\midrule
AM    & 0.82 &0.97&4.02&2.72&3.41&2.99&1.63&1.74\\
POMO &0.00 & 0.18&0.41&0.84&0.01&1.84&0.83&0.31\\
LEHD &0.22 &0.32&0.12&0.38&0.01&2.31&1.64&0.33\\
GNNGLS &0.04&1.57&0.73&0.57&0.46&0.20&0.00&0.76\\
NeuralGLS &0.82&1.77&0.03&0.00&0.00&0.36&0.00&0.00\\
GLS &0.00&0.50&0.00&0.02&0.01&3.28&0.67&0.31\\
EBGLS &0.00&0.01&0.20&0.02&0.01&1.91&0.67&0.31\\
KGLS &0.00&0.01&0.00&0.06&0.02 &2.07&0.67&0.31\\
EOH    & 0.00  & 0.01  & 0.00 & 0.02  & 0.01 & 2.27&0.67 &0.31\\
\textbf{FWA + MLLM + visual information} & \textbf{0.00} & \textbf{0.00} & \textbf{0.00} & \textbf{0.00} & \textbf{0.00} & \textbf{0.00}   & 0.23  & 0.15\\
\textbf{FWA + MLLM + w/o visual information} & \textbf{0.00} & \textbf{0.00} & \textbf{0.00} & \textbf{0.00} & \textbf{0.00}   & \textbf{0.00} & \textbf{0.00}   & \textbf{0.00}  \\
\bottomrule
\end{tabular}}
\end{table*}

Some TSP instances with optimal paths (for easy visualization and comparison) were selected for testing, including pr76, kroA100, kroC100, kroD100, rd100, eil101, eil51, st70.  Our algorithm is compared with algorithms with learning modules (e.g., AM\cite{16}, POMO\cite{17}, LEHD\cite{18}, etc.), and heuristic algorithms (e.g., GLS\cite{19}, GNNGLS\cite{22}, EBGLS\cite{20}, KGLS\cite{21}, NeuralGLS\cite{23}, EOH\cite{4}, etc.).

\begin{figure*} % 星号用于创建跨双栏图像
  \centering
  \begin{subfigure}{\textwidth}
    \includegraphics[width=\textwidth]{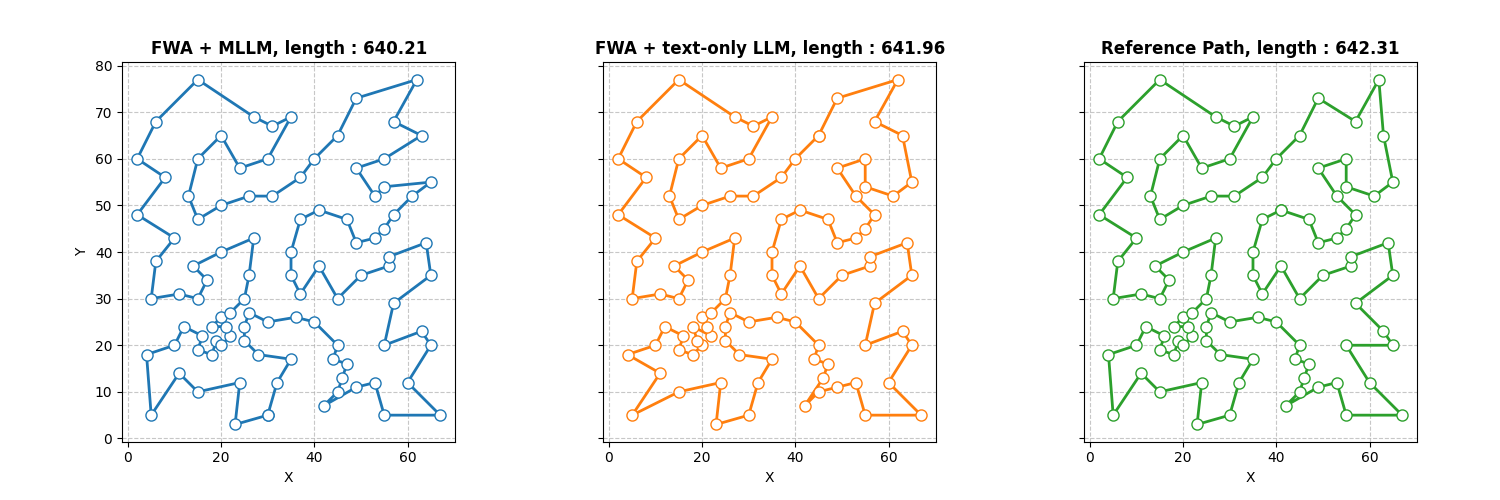}
    \caption{}
    \label{fig:eil101}
  \end{subfigure}
  \hfill
  \begin{subfigure}{\textwidth}
    \includegraphics[width=\textwidth]{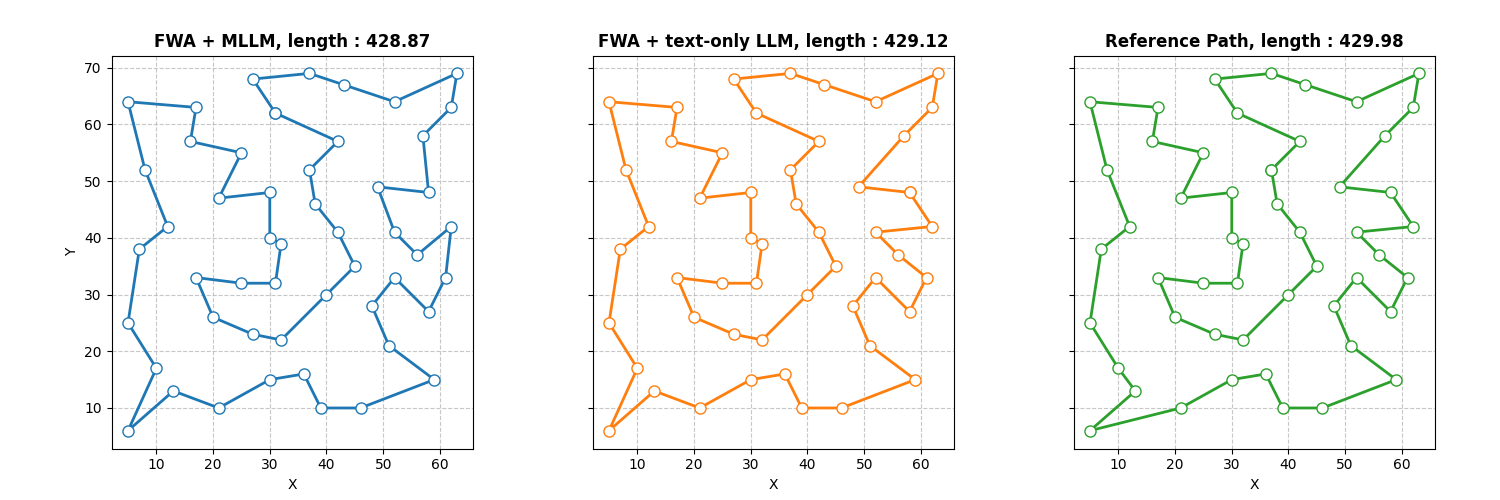}
    \caption{}
    \label{fig:eil51}
  \end{subfigure}
  \hfill
  \begin{subfigure}{\textwidth}
    \includegraphics[width=\textwidth]{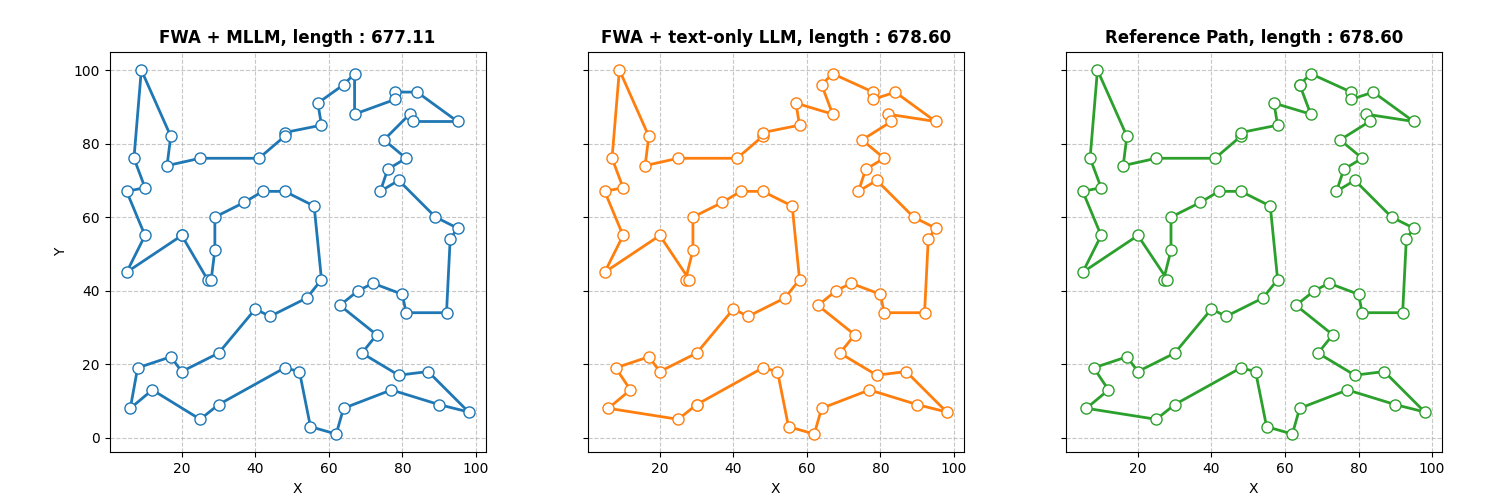}
    \caption{}
    \label{fig:st70}
  \end{subfigure}
  
  \caption{Each row represents a comparison of three optimal paths on a TSP instance, corresponding to fwa + MLLM + visual information, fwa + MLLM + w/o visual information, reference path in TSPLIB. (a)  eil101 (b)  eil51 (c) st70}
  \label{fig:tsp}
\end{figure*}

The results of the subset problems is shown in Table \ref{table:tsp_results}. Our algorithm consistently \textbf{outperforms all leading TSP heuristics/algorithms }across diverse TSPLib instances. It easily can be seen that the introduction of visual modalities in the optimization process does not guarantee to improve the performance of the overall framework, which is against our intuition. On eil51 and st70, our framework without visual information can do better than that with visual information. It is also worth mentioning that we are able to find different optimal paths different from those in TSPLIB (note that the comparisons in Table I are made with distances rounded, but if under floating-point arithmetic, our paths turn out to be even better than the paths in TSPLIB on eil101, eil51 and st70, see Figure \ref{fig:tsp}). \textbf{The better solutions in the floating-point sense we get in these instances are all derived from the FWA generated when visual information is given} (although the rounding results may not be as good as the that without visual information, as shown in Table \ref{table:tsp_results}).

\begin{figure*}[htbp]
\centering
\begin{minipage}[t]{0.48\textwidth}
\centering
\includegraphics[width=8cm]{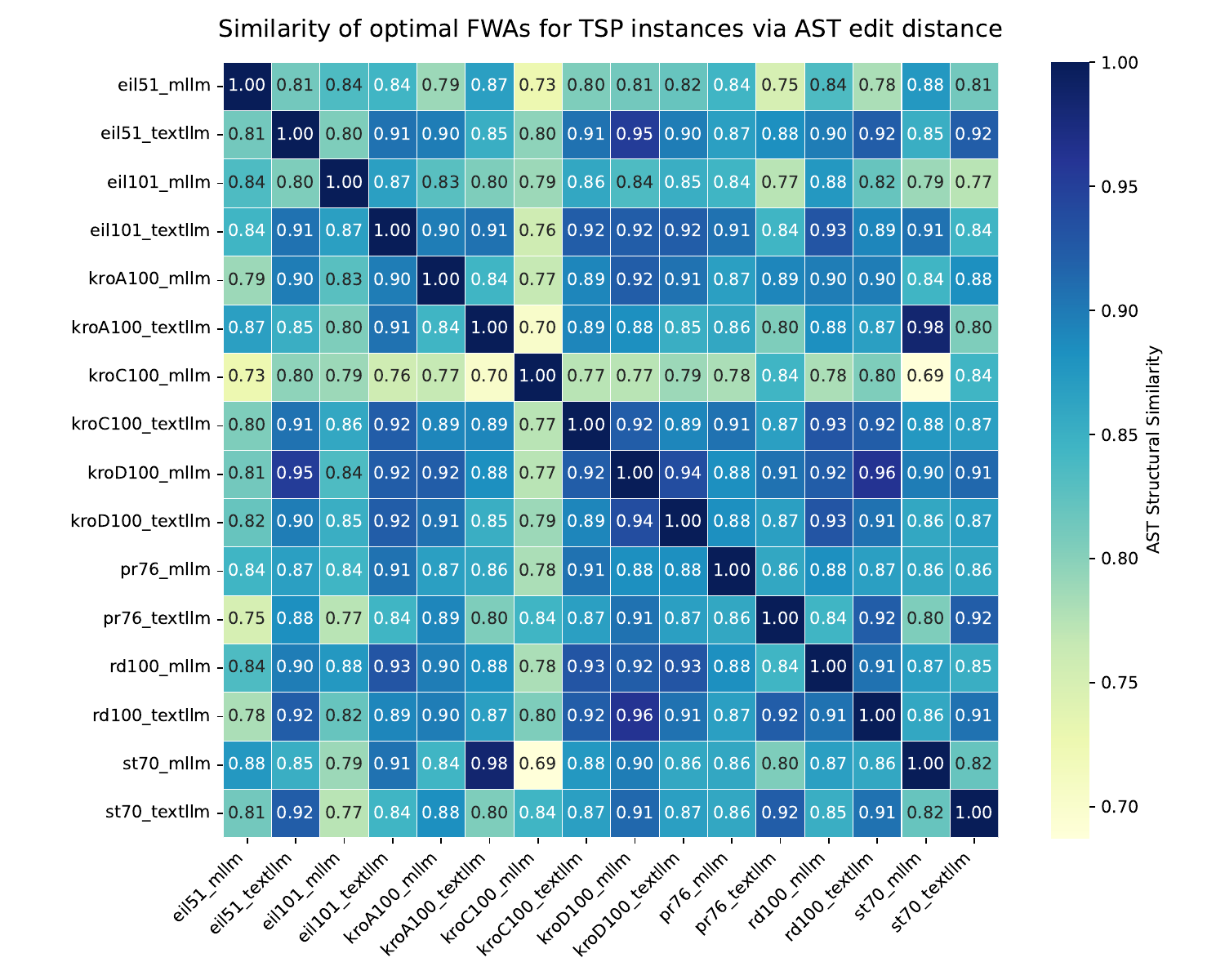}
\end{minipage}
\begin{minipage}[t]{0.48\textwidth}
\centering
\includegraphics[width=8cm]{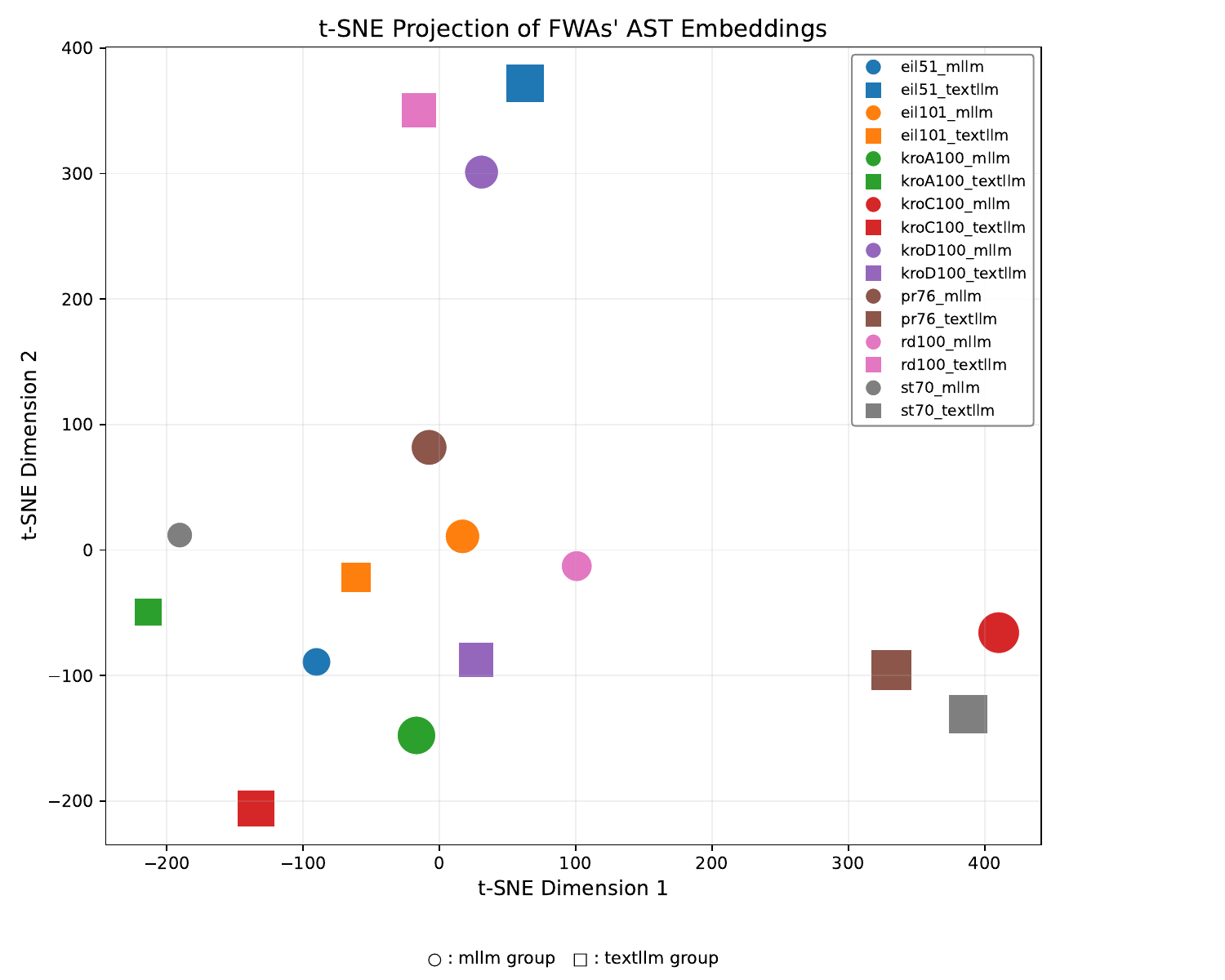}
\end{minipage}
\begin{minipage}[t]{0.48\textwidth}
\centering
\includegraphics[width=8cm]{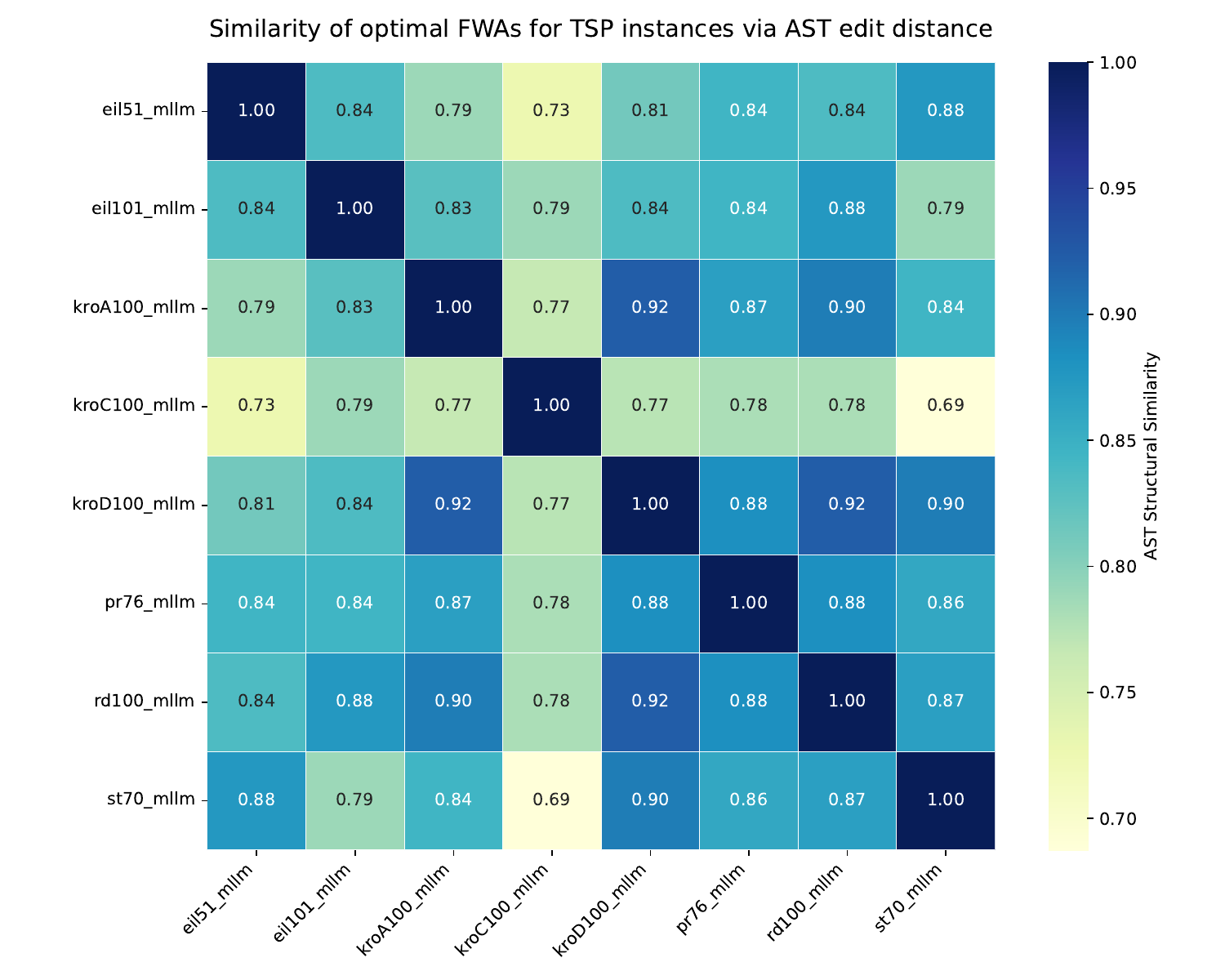}
\end{minipage}
\begin{minipage}[t]{0.48\textwidth}
\centering
\includegraphics[width=8cm]{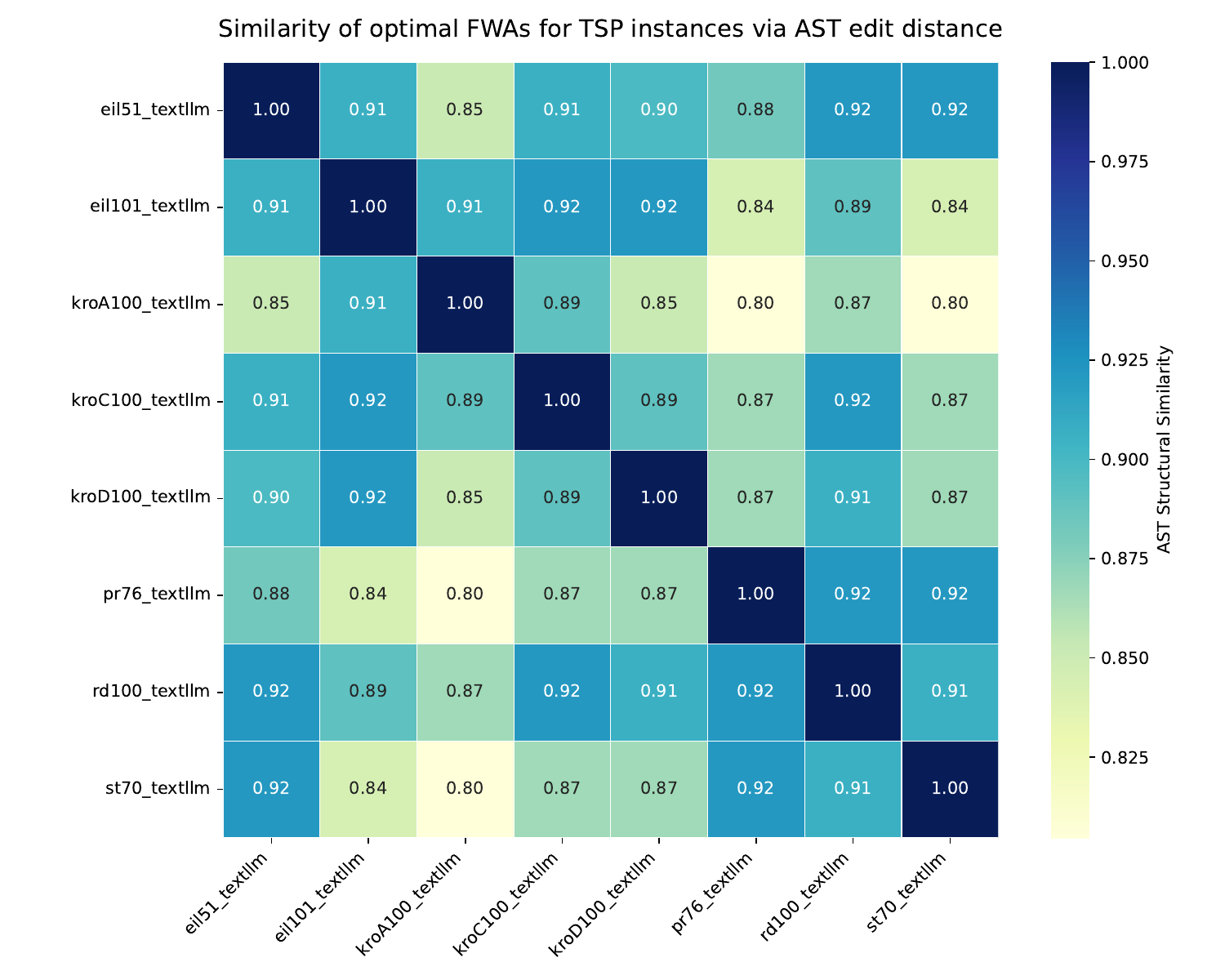}
\end{minipage}
\caption{Similarity measure matrix(upper Left: all FWAs; lower left: FWAs produced by MLLM + visual information; lower right: FWAs produced by MLLM + w/o visual information) and t-SNE visualization(Upper Right) of optimal FWAs for different TSP instances based on their abstract syntax trees}
\label{fig:tsp_2}
\end{figure*}

In Figure \ref{fig:tsp_2}, we selected all optimal FWAs produced in different TSP instances for further analysis. As each problem instance corresponds to with or without visual information two situations, 16 distinct code implementations were subjected to similarity quantification and visualization. To mitigate the influence of superficial syntactic variations (e.g., identifier naming), each code sample underwent Abstract Syntax Tree (AST) analysis first. The upper-left panel visualizes pairwise similarity computed via Jaccard distance between AST node sets, while the upper-right panel projects AST representations into a 2D embedding space using a text-embedding model(doubao-embedding-large-text-250515) followed by t-SNE dimensionality reduction. The size of the points in the upper-right is proportional to the length of the optimal FWA, and it can be seen that the length of the FWA is relatively uniform. The matrices in lower left panel and lower right panel are extracted from the whole similarity matrix to show the effect of visual information. In Fig.\ref{fig:tsp_2}, FWAs in the lower-left panel are generated with visual information (denoted as \textit{instance\_mllm}), whereas FWAs in the lower-right panel are generated without visual inputs (denoted as \textit{instance\_textllm}). Here are some key points:\\
\begin{enumerate}
    \item \textbf{FWAs generated by MLLM without visual information exhibit higher structural homogeneity in TSP}. FWAs designed solely using textual information demonstrate high structural convergence across diverse TSP problem instances. This indicates that the absence of visual input constrains the algorithm design process, leading to the generation of near-identical solution structures regardless of instance-specific geometric properties. The average cross-instance similarity with visual information is 0.8848±0.03 (mean±SD), significantly exceeding that without visual information (0.8325±0.06) with p-value $<$ 0.05. confirming limited structural adaptation to instance geometry, which also means that the visual information helps MLLM to capture the unique geometric features of the paths obtained from the FWA in each instance, thus generating more customized operators. 
    \item \textbf{Visual information leads to significant but non-fundamental modifications on the same instance in TSP}. For the same TSP instance, there is a measurable similarity  between the FWA generated with visual information and the FWA generated without visual information, but the similarity is still significantly smaller than the ultra-high similarity among FWA with p-value $<$ 0.05. This suggests that the visual information is used to tune or optimize the general logic for FWA generation.
\end{enumerate}

\subsection{Electronic Design Automation Problem}
The placement of components in integrated circuits constitutes a pivotal stage in physical design, where the primary objective is to arrange millions of functional units without overlaps while minimizing the total  wirelength(here we use the DreamPlace framework to explore the results of wirelength, but other indicators can also be included later).This task directly governs chip performance metrics such as signal delay, power consumption, and manufacturing yield, with suboptimal placements leading to catastrophic timing violations or irresolvable routing congestion. The problem’s complexity escalates exponentially with modern IC scaling—a design with $10^5$ components introduces over $10^9$ pairwise non-overlapping constraints, rendering conventional optimization methods intractable due to the curse of dimensionality. 

The \texttt{DreamPlace}\cite{12,13} framework revolutionizes IC placement by reformulating discrete component positioning as a continuous optimization problem. Its core algorithm minimizes a composite objective function combining wirelength and density penalties:

\begin{equation}
\label{eq:dreamplace_obj}
\min_{\mathbf{x,y\in R^n}} f(x) =\sum_{e\in E}w_{e}W(e;x,y) + \lambda D(\mathbf{x,y}), \quad \lambda > 0
\end{equation}

where $\mathbf{x,y} \in \mathbb{R}^{n}$ represents component coordinates, $E$ denotes netlist connections, and $D(\mathbf{x,y})$ enforces overlap constraints via a smoothed electrostatic analogy. $W(e;x,y)$ is half-parameter wirelength model. Nonlinear placement adopts a nonlinear differentiable approximation of $W(e;x,y)$. A widely-used approximation is the weighted-average (WA) model shown below:
\begin{equation}
    \widetilde{W}_x(e,\gamma,x,y)=\frac{\sum_{i\in e}{x_ie^{\frac{x_i}{\gamma}}}}{\sum_{i\in e}{e^{\frac{x_i}{\gamma}}}}+\frac{\sum_{i\in e}{x_ie^{-\frac{x_i}{\gamma}}}}{\sum_{i\in e}{e^{-\frac{x_i}{\gamma}}}}
\end{equation}
\begin{equation}
    \widetilde{W}(e,\gamma,x,y)=\widetilde{W}_x(e,\gamma,x,y)+\widetilde{W}_y(e,\gamma,x,y)
\end{equation}
The framework employs conjugate gradient descent with Armijo line search, accelerated by GPU-based sparse matrix operations. 

Despite its computational merits, \texttt{DreamPlace}'s reliance on the Barzilai-Borwein (BB) step-size\cite{24} heuristic introduces critical instability. The BBSTEP heuristic is shown in Eq.\ref{eq:bbstep}.

\begin{equation}
\label{eq:bbstep}
\alpha_k = \frac{\mathbf{s}_{k-1}^\top \mathbf{y}_{k-1}}{\mathbf{y}_{k-1}^\top \mathbf{y}_{k-1}}, \quad 
\begin{cases}
\mathbf{s}_{k-1} = \mathbf{x}_k - \mathbf{x}_{k-1} \\
\mathbf{y}_{k-1} = \nabla f(\mathbf{x}_k) - \nabla f(\mathbf{x}_{k-1})
\end{cases}
\end{equation}

\begin{equation}
    \widetilde{x}_{k+1} = x_k-\alpha_k  \nabla f(\mathbf{x}_k) 
\end{equation}

\begin{equation}
    \theta_{k+1}=\frac{1+\sqrt{1+4\theta_k^2}}{2}
\end{equation}

\begin{equation}
    {x}_{k+1} = \widetilde{x}_{k+1} +  \frac{\theta_k - 1}{\theta_{k+1}} (\widetilde{x}_{k+1}-\widetilde{x}_{k})
\end{equation}
While the BBSTEP heuristic provides empirically viable step-sizes in moderate-curvature regions, its one-shot estimation proves inadequate for non-convex EDA objectives where local geometry varies drastically across the parameter space. Inspired by the FWA, we improve the step size design and get better result.

Different from the TSP problem, instead of adopting FWA as the overall framework in the EDA task, we absorb FWA's optimization idea and design heuristic optimization rules, specifically, we need the multimodal large language model to design the CP as shown below, keep the input unchanged and return a tensor output. 
\begin{tcolorbox}[title=Critical part in DreamPlace for EDA]

\begin{verbatim}
with torch.no_grad():
    s_k = (v_k - v_k_1)
    y_k = (g_k - g_k_1)
    bb_long_step_size = (s_k.dot(s_k) / 
        torch.sum(s_k * y_k)).data
    bb_short_step_size = (s_k.dot(y_k)/ 
        y_k.dot(y_k)).data
    lip_step_size = (s_k.norm(p=2) /
        y_k.norm(p=2)).data 
    base_step = bb_short_step_size if 
        bb_short_step_size > 0 else 
        min(lip_step_size, alpha_k)

#######Critical Part is here#######
step_size = firework_step_redesign(
    v_k=v_k,
    g_k=g_k,
    fn=obj_fn,
    base_step=base_step,
)


u_kp1 = v_k - step_size*g_k
v_kp1.data.copy_(u_kp1 + coef*
    (u_kp1-u_k))
constraint_fn(v_kp1)
\end{verbatim}
\end{tcolorbox}

\begin{tcolorbox}[title=Initial step design for our Framework]

\begin{verbatim}
import torch

def firework_step_redesign(v_k, g_k, 
    fn, base_step):
    '''
    params:
    v_k:current placement
    g_k:current gradient 
    fn: evaluation tool for a solution
    base_step: \alpha_k in eq (2)
    '''
    return base_step
\end{verbatim}
\end{tcolorbox}
Similarly we will design two operations, including mutation and crossover, which are almost consistent with TSP in the textual information part except for a important priori that returning "base\_step" directly is a good baseline. When visual information is provided, we adopt the visualization of placement which is provided in Dreamplace as the supplementary input. You can refer to Fig.\ref{fig:EDA} for details. The prompts for the two operations are shown below.

\begin{figure}
    \centering
    \includegraphics[width=0.7\linewidth]{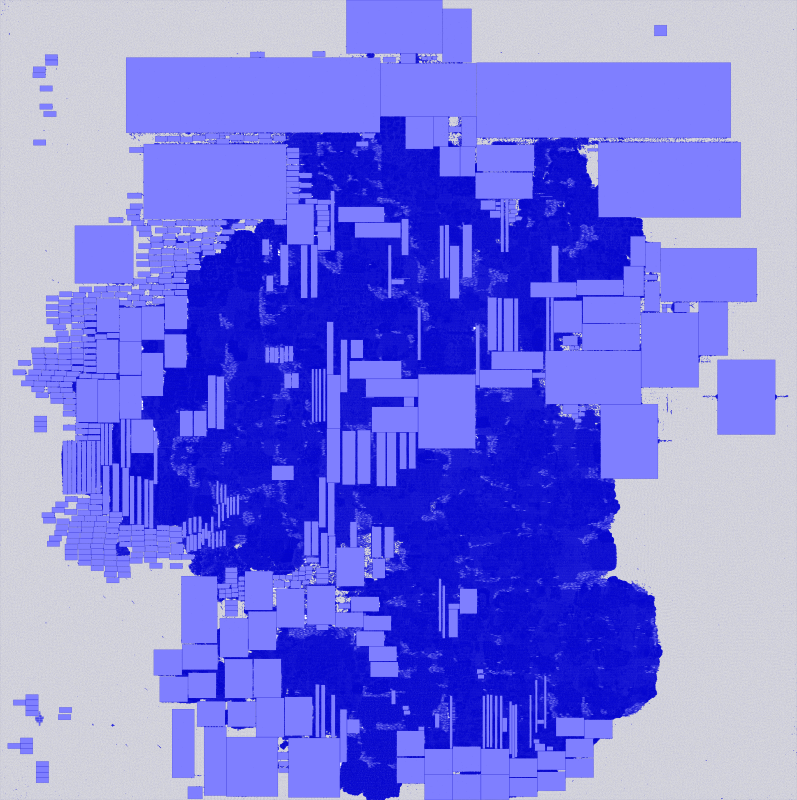}
    \caption{Visualization of EDA task layout results}
    \label{fig:EDA}
\end{figure}

\begin{mutationbox_eda}
\begin{lstlisting}
You are an expert in integrated circuit placement optimization using Fireworks Algorithm to enhance DREAMPlace's step size controller. Generate an improved FWA code using mutation based on the following information:
Current Performance Data
    1. Total wHPWL: {parent[0]}
    2. Current FWA code to control the step: {parent[1]} 
    3. Current placement graph: refer to the picture I send you(Only provided for when visual information is given)
Input of the code:
    1.v_k: current placement, on cuda
    2.g_K: current placement's gradient, on cuda
    3.fn : a function to get the result (a real number) of the placement x by calling fn(x)
    4.base_step: a good initial step you can refer
Ouput of the code:
    A better step for the current placement v_k (should be the same dtype/device/shape as base_step)
Modification Requirements:
    1. Maintain functions' original input formats
    2. Be creative and brave to design the function
    3. Return ONLY the complete correct Python code with necessary imports, no explanations
An important prior:
    1. A good baseline is just reutrn base_step, so you should try your best to conquer it
\end{lstlisting}
\end{mutationbox_eda}

\begin{crossoverbox_eda}
\begin{lstlisting}
You are an expert in integrated circuit placement optimization using Fireworks Algorithm to enhance DREAMPlace's step size controller. Please perform crossover on the following two high-performing codes:
Current Performance Data
    parent 1:
    1. Total wHPWL: {parents[0][0]}
    2. Current FWA code to control the step: {parents[0][1]} 
    3. Current placement graph: refer to the picture I send you(Only provided for when visual information is given)
    parent 2:
    1. Total wHPWL: {parents[1][0]}
    2. Current FWA code to control the step: {parents[1][1]} 
    3. Current placement graph: refer to the picture I send you(Only provided for when visual information is given)
Input of the code:
    1.v_k: current placement, on cuda
    2.g_K: current placement's gradient, on cuda
    3.fn: a function to get the result (a real number) of the placement x by calling fn(x)
    4.base_step: a good initial step you can refer
Ouput of the code:
    A better step for the current placement v_k (should be the same dtype/device/shape as base_step)
Modification Requirements:
    1. Maintain functions' original input formats
    2. Be creative and brave to design the function
    3. Return ONLY the complete correct Python code with necessary imports, no explanations
An important prior:
    1. A good baseline is just reutrn base_step, so you should try your best to conquer it
\end{lstlisting}
\end{crossoverbox_eda}

We adopted the I/O-free ISPD2005 benchmark offering greater layout flexibility, comprising eight sub-problems. The minimum dimension of these 8 optimization instances is 570,000 and the maximum is 10 million. As we say at the beginning, our demand for computing resources is extremely low. Each case was executed on \textbf{a single 12GB NVIDIA TITAN GPU }without acceleration following a case-specific algorithm design methodology, as individual instances potentially required distinct step-size adjustment strategies. Our evolutionary framework generated two alternative strategies per iteration through mutation and crossover operations, configured with a maximum of 200 generations and maintained a population size of 5 (managed via greedy selection). The comparative baselines include the original Dreamplace framework and EOAGP\cite{11}, which exclusively leverages LLMs' text generation capabilities to improve the whole Dreamplace in several aspects, including initialization, preconditioner, and optimizer. EOAGP uses \textbf{a distributed GPU cluster comprising 60 NVIDIA RTX 2080 Ti and 150 RTX 3090 GPUs for parallel acceleration}, much more than ours.

\begin{figure}[htbp]
    \centering
    \begin{subfigure}{0.6\linewidth}
        \centering
        \includegraphics[width=\linewidth]{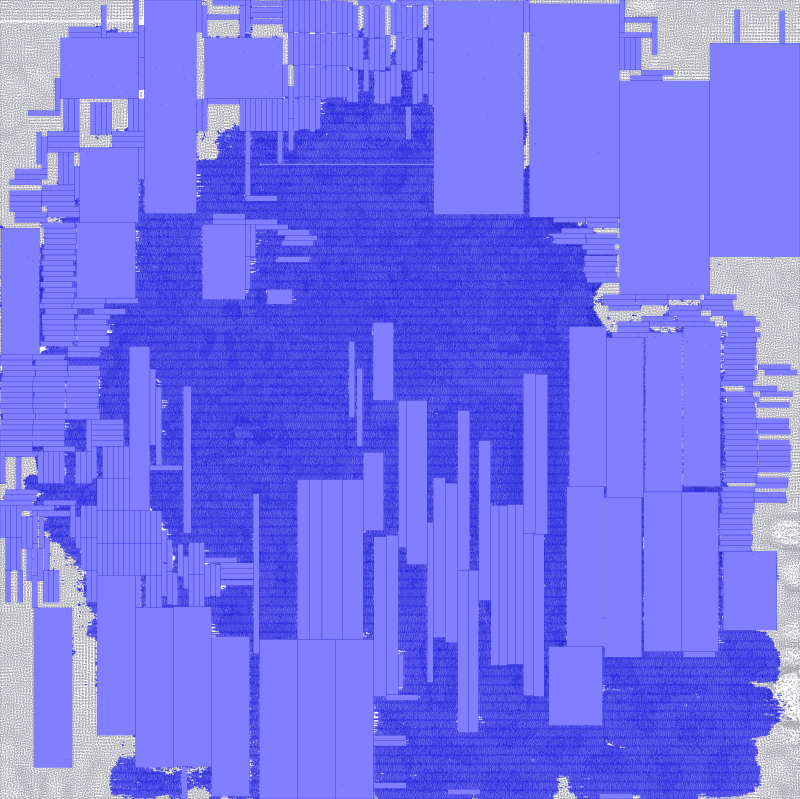}
        \caption{The old  solution for adaptec1 of I/O-free ISPD2005 benchmark}
        \label{fig:image1}
    \end{subfigure}
    \vskip 10pt % 两张图片间的垂直间距（可选）
    \begin{subfigure}{0.6\linewidth}
        \centering
        \includegraphics[width=\linewidth]{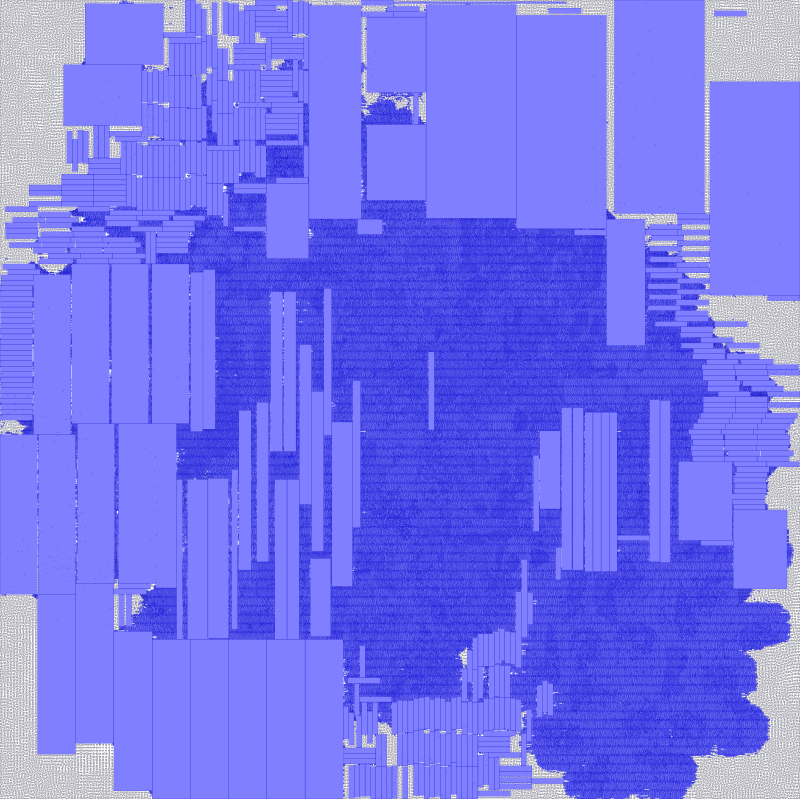}
        \caption{The new better solution for adaptec1 of I/O-free ISPD2005 benchmark}
        \label{fig:image2}
    \end{subfigure}
    \caption{The comparsion between the current best solution and our new solution on adaptec1 instance (14.98\% better than before)}
    \label{fig:results_eda_a1}
\end{figure}

Please see Table \ref{tab:eda_results} for specific experimental results. Notably, all runtime metrics are normalized against the Dreamplace framework's execution time (set as baseline 1.0 for each design case), providing a consistent evaluation criterion. Based on the comprehensive experimental evaluation, our methodology demonstrates compelling performance (all better than Dreamplace), particularly in achieving superior wirelength minimization and computational efficiency.  Our approach achieved the lowest wHPWL values across 6 of 8 benchmarks, with particularly significant reductions in large-scale designs:

Our approach is useful for both small and large scale instances. Comparing with other methods, for adaptec1, we attained a wHPWL of 61.81 (vs. 68.44 for EOAGP and 72.71 for Dreamplace). For bigblue4, we achieved 586.57 wHPWL (outperforming EOAGP’s 587.77 and Dreamplace’s 612.45). From the final optimal code generated, they tend to call the "fn" function multiple times to compute the performance of the current candidate step, which can introduce additional computation time. The observed variations in runtime reflect deliberate quality-efficiency trade-offs, where marginal time investments often bring substantial wHPWL improvements. We show a comparison plot of the results on the adaptec1 problem instance, where we can see that the original scheme (Fig.\ref{fig:image1}) has more irregular gray areas and quite a lot of fragmented distribution of small blue layout blocks, which is mitigated and improved in our solution (Fig.\ref{fig:image2}). Comparing within the FWAs, it can be seen that both cases(with and without visual optimization information) performed better on half of the tasks, respectively, but the former winners with visual optimization information were concentrated in high-dimensional instances (a3, b2, b3, b4). This suggests that the \textbf{introduction of visual modal information is useful for optimizer design in high-dimensional complex problems}. At the same time, we can see that the introduction of visual modal optimization information \textbf{does not guarantee better algorithm performance}(a1, a2, a4, b1), which is again against our instincts.

\begin{figure*}[htbp]
\centering
\begin{minipage}[t]{0.48\textwidth}
\centering
\includegraphics[width=8cm]{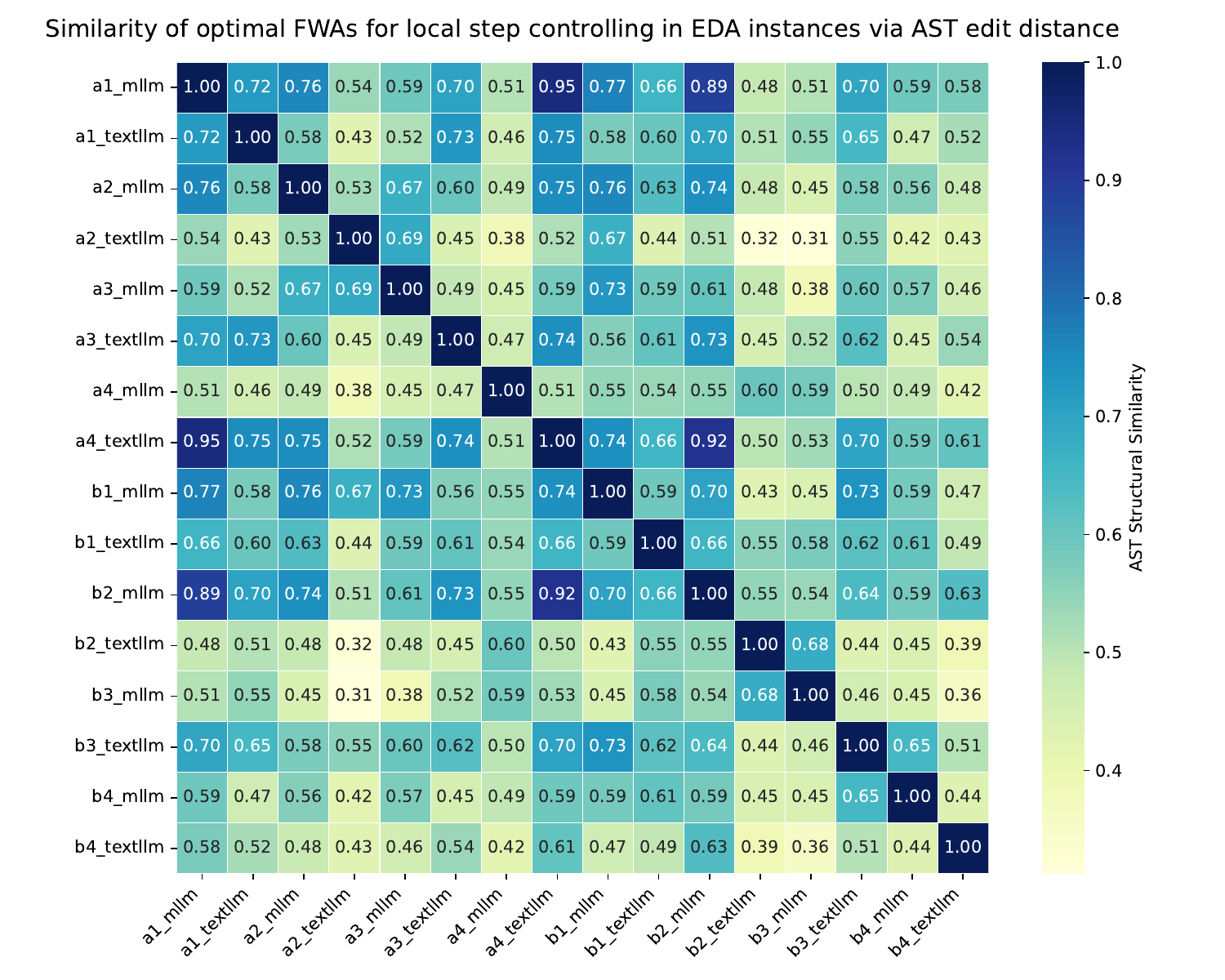}
\end{minipage}
\begin{minipage}[t]{0.48\textwidth}
\centering
\includegraphics[width=8cm]{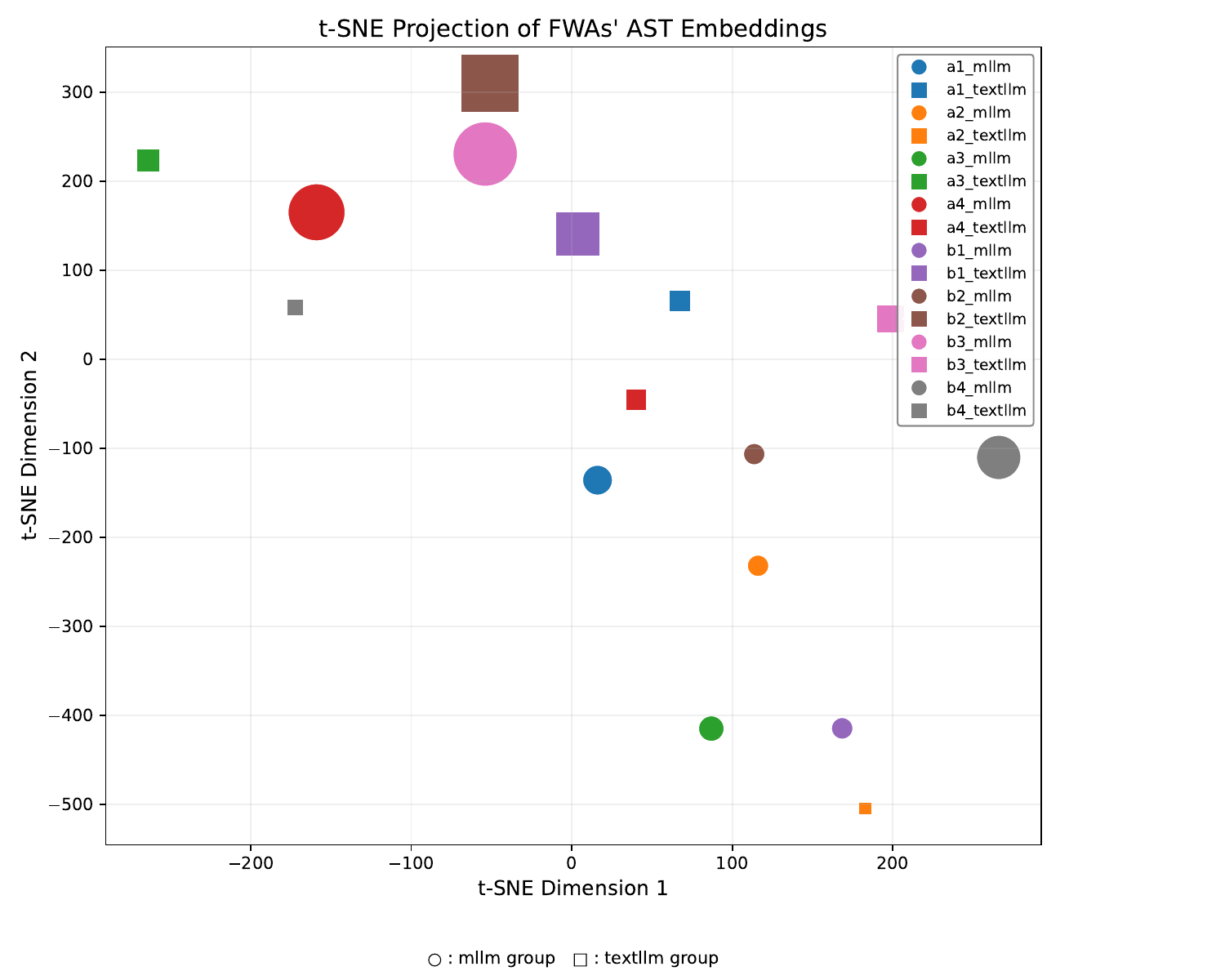}
\end{minipage}
\begin{minipage}[t]{0.48\textwidth}
\centering
\includegraphics[width=8cm]{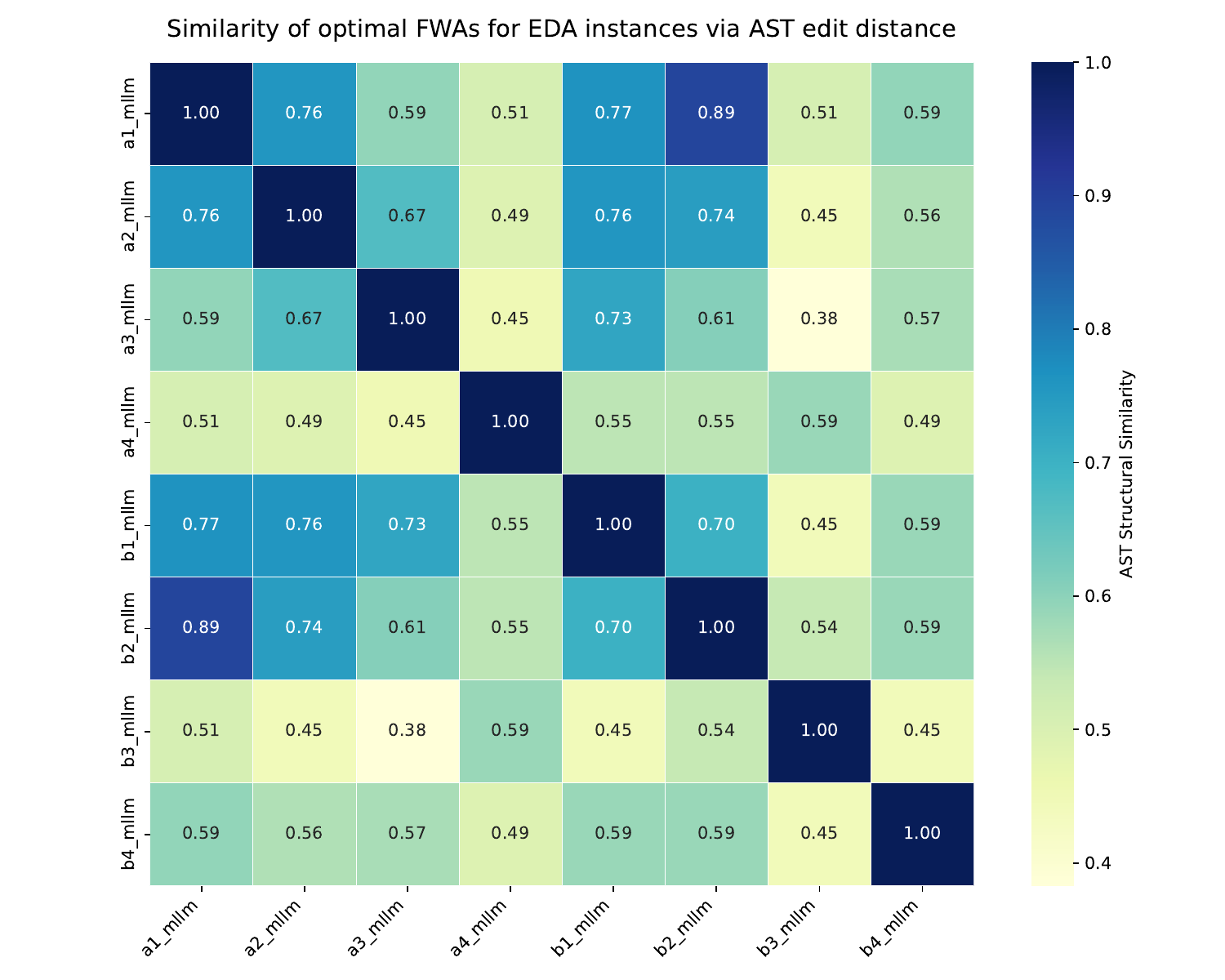}
\end{minipage}
\begin{minipage}[t]{0.48\textwidth}
\centering
\includegraphics[width=8cm]{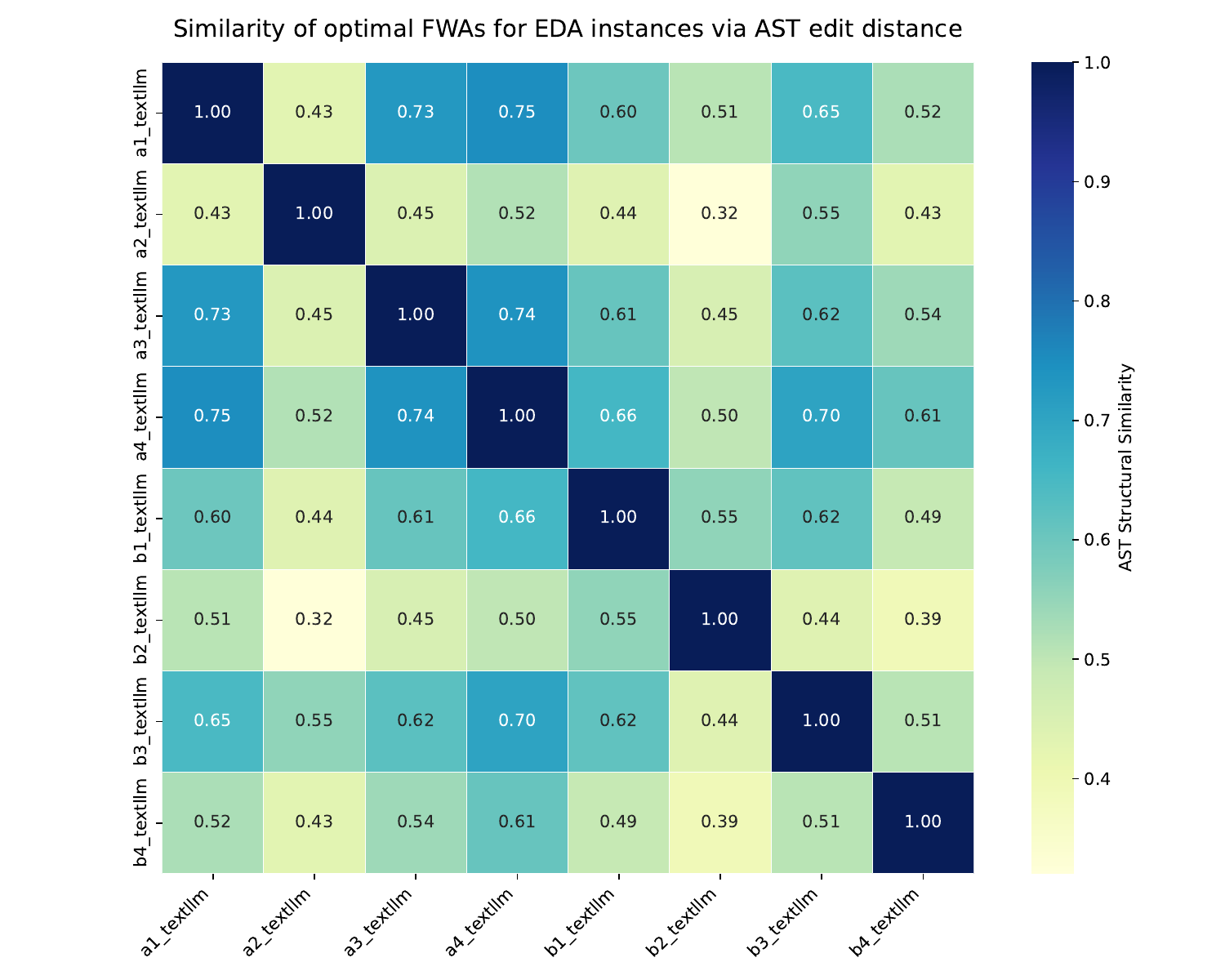}
\end{minipage}
\caption{Similarity measure matrix((upper left: all FWAs; lower left: FWAs produced by MLLM; lower right: FWAs produced by TextLLM) and t-SNE visualization(Upper Right)) and t-SNE visualization(uper right) of optimal FWAs for different EDA instances based on their abstract syntax trees}
\label{fig:eda_analysis}
\end{figure*}

With the help of AST, we also performed a similarity analysis on the optimal FWAs  generated by MLLM in EDA tasks, hoping to further understand the behavior of MLLM in the process of algorithm generation. Here are some key points:
\begin{enumerate}
    \item From the upper-right panel, we can see that the distribution of optimal FWAs code lengths on different task instances in EDA is very uneven, but the optimal FWA code lengths generated in EDA tasks like a1, a2, etc., are shorter and closer together in the 2-dimensional embedding.
    \item In EDA task, the calculated p-values no longer support the statistically significant conclusion that "the use of visual modality optimization information significantly reduces the similarity of MLLM-generated FWAs" got in TSP experiment. In the TSP experiments, due to the low dimensionality, the visual information clearly reflects the difference between the different instances, including city distribution, current path characteristics, etc. However, the visual information in the EDA task( is more similar, as the problem dimensions are very large, MLLM cannot accurately capture the condition of the vast majority of individuals, and the layout diagrams of different instances may show similar topology at the pixel level (e.g., many pieces of the layout space, dispersion of layout results, etc.), which results in the visual information instead amplifying the surface commonalities, and thus the similarity of the FWAs generated with visual information is elevated.
\end{enumerate}

We focus on the point of optimizer's step size adjustment, with a priori that the core of the whole Dreamplace is on the underlying optimizer, and the optimizer is very sensitive to the optimizer's step size. And the experiment results support our priori.

\begin{table*}[htbp]
    \centering
    \caption{Comparison of Different Methods on I/O free ISPD2005 benchmark}
    \label{tab:eda_results}
    \renewcommand{\arraystretch}{1.2} % 增加行高
     \resizebox{\textwidth}{!}{
    \begin{tabular}{|l|c|c|c|c|c|c|c|c|}
        \hline
        \multirow{2}{*}{Design Case} & \multicolumn{2}{c|}{Dreamplace w/ BB Method} & \multicolumn{2}{c|}{EOAGP} & \multicolumn{2}{c|}{FWA + MLLM + visual information} & \multicolumn{2}{c|}{FWA + MLLM + w/o visual information}  \\
        \cline{2-9}
         & wHPWL & Runtime & wHPWL & Runtime & wHPWL & Runtime & wHPWL & Runtime \\
        \hline
        adaptec1(dim $\approx$ 570k) & 72.71 & 1.0 & 68.44 & 0.88 & 61.82 & 1.51  & \textbf{61.81} & 2.52\\
        \hline
        adaptec2(dim $\approx$ 780k) & 82.46 & 1.0 & 80.99 & 1.08 & 73.58 & 1.11 &\textbf{72.28} & 1.30 \\
        \hline
        adaptec3(dim $\approx$ 2.77M) & 133.61 & 1.0 & 127.60 & 0.98 & \textbf{127.31} & 1.28 &130.00 & 1.36 \\
        \hline
        adaptec4(dim $\approx$ 3.75M) & 132.67 & 1.0 & 125.96 & 0.98 & 125.92 & 1.02 & \textbf{125.81} &1.10 \\
        \hline
        bigblue1(dim $\approx$ 1.05M) & 82.56 & 1.0 & 80.90 & 1.01 & 80.88 & 0.98 & \textbf{80.72} & 1.13 \\
        \hline
        bigblue2(dim $\approx$ 2.97M) & 98.75 & 1.0 & \textbf{91.98} & 1.01 & 93.94 & 1.03  & 94.77 & 0.98\\
        \hline
        bigblue3(dim $\approx$ 4.08M) & 287.13 & 1.0 & \textbf{252.12} & 0.88 & 265.85 & 1.34  & 272.75 & 1.19\\
        \hline
        bigblue4(dim $\approx$ 10.09M) & 612.45 & 1.0 & 587.77 & 1.15 & \textbf{586.57} & 1.07 & 602.03 & 1.49  \\
        \hline
    \end{tabular}}
\end{table*}

\section{Conclusion}
In this work, we propose a framework that leverages MLLM to enhance the FWA for optimizing complex problems. And it's obvious that this framework can be quickly migrated to other swarm intelligence optimization algorithms or even first-order optimizers. Primarily, visual information is integrated to augment MLLM's comprehension of both the overarching task and real-time optimization outcomes. Crucially, we introduce the concept of CP, which hybridizes direct evolutionary optimizers (tested on TSP) with FWA's idea-driven evolution (tested on EDA). This innovation significantly broadens the applicability scope of population-based swarm intelligence optimization algorithms like FWA. Experiments results also demonstrate our framework's superiority:
\begin{enumerate}
    \item For TSP, it outperforms all leading TSP heuristics/algorithms, and give better paths in floating-point sense.
    \item For EDA, our framework delivers SOTA performance on 6 out of 8 benchmarks using very limited computational resources, further attesting to its efficacy.
\end{enumerate}

We also analyzed the design behavior of MLLM in our framework when visual information is given:
\begin{enumerate}
    \item \textbf{At the performance level}. The introduction of visual information in the optimization process does not guarantee to improve the performance of the overall framework. But it's still worth a try for high-dimensional, challenging tasks.
    \item \textbf{At algorithmic similarity level}. If the visual information can clearly characterize the current task and the result (in the case of TSP), the heterogeneity of the optimization algorithms generated by MLLM increases; however, if the visual information is not well characterized (in the case of EDA), MLLM instead amplifies the surface commonalities, leading to an increase in the homogeneity of optimization algorithms between different instances.
\end{enumerate}
This dual-domain advancement—breaking optimization records while maintaining resource efficiency—empirically validates our framework's capability to transcend traditional FWA limitations and generalize across combinatorial and physical design challenges.

\section*{acknowledgement}
This work is supported by the National Natural Science
Foundation of China (Grant No. 62076010), and partially
supported by the National Key R\&D of China (Grant
\#2022YFF0800601).

\vspace{12pt}
\end{document}